\documentclass[10pt,twocolumn,letterpaper]{article}

\usepackage{cvpr}
\usepackage{times}
\usepackage{epsfig}
\usepackage{graphicx}
\usepackage{amsmath}
\usepackage{amssymb}

\usepackage{algorithm,algorithmic}
\usepackage{breqn}
\usepackage{booktabs}
\usepackage{multirow}
\usepackage{subcaption}

\usepackage[pagebackref=true,breaklinks=true,letterpaper=true,colorlinks,bookmarks=false]{hyperref}

\cvprfinalcopy 

\def\cvprPaperID{5031} 

\ifcvprfinal\fi
\begin{document}

\title{Deep Bayesian Active Learning for Multiple Correct Outputs}

\author{Khaled Jedoui, Ranjay Krishna, Michael Bernstein, Li Fei-Fei\\
Stanford University\\
{\tt\small {\{thekej, ranjaykrishna, msb, feifeili\}}@cs.stanford.edu}
}
\maketitle

\begin{abstract}

Typical active learning strategies are designed for tasks, such as classification, with the assumption that the output space is mutually exclusive. The assumption that these tasks always have exactly one correct answer has resulted in the creation of numerous uncertainty-based measurements, such as entropy and least confidence, which operate over a model's outputs. Unfortunately, many real-world vision tasks, like visual question answering and image captioning, have multiple correct answers, causing these measurements to overestimate uncertainty and sometimes perform worse than a random sampling baseline.
In this paper, we propose a new paradigm that estimates uncertainty in the model's internal hidden space instead of the model's output space.
We specifically study a manifestation of this problem for visual question answer generation (VQA), where the aim is not to classify the correct answer but to produce a natural language answer, given an image and a question. Our method overcomes the paraphrastic nature of language. It requires a semantic space that structures the model’s output concepts and that enables the usage of techniques like dropout-based  Bayesian uncertainty. We build a visual-semantic space that embeds paraphrases close together for any existing VQA model.
We empirically show state-of-art active learning results on the task of VQA on two datasets, being $5$ times more cost-efficient on Visual Genome and $3$ times more cost-efficient on VQA 2.0.
\end{abstract}

\section{Introduction}

\begin{figure}[t]
\begin{center}
\includegraphics[width=\linewidth]{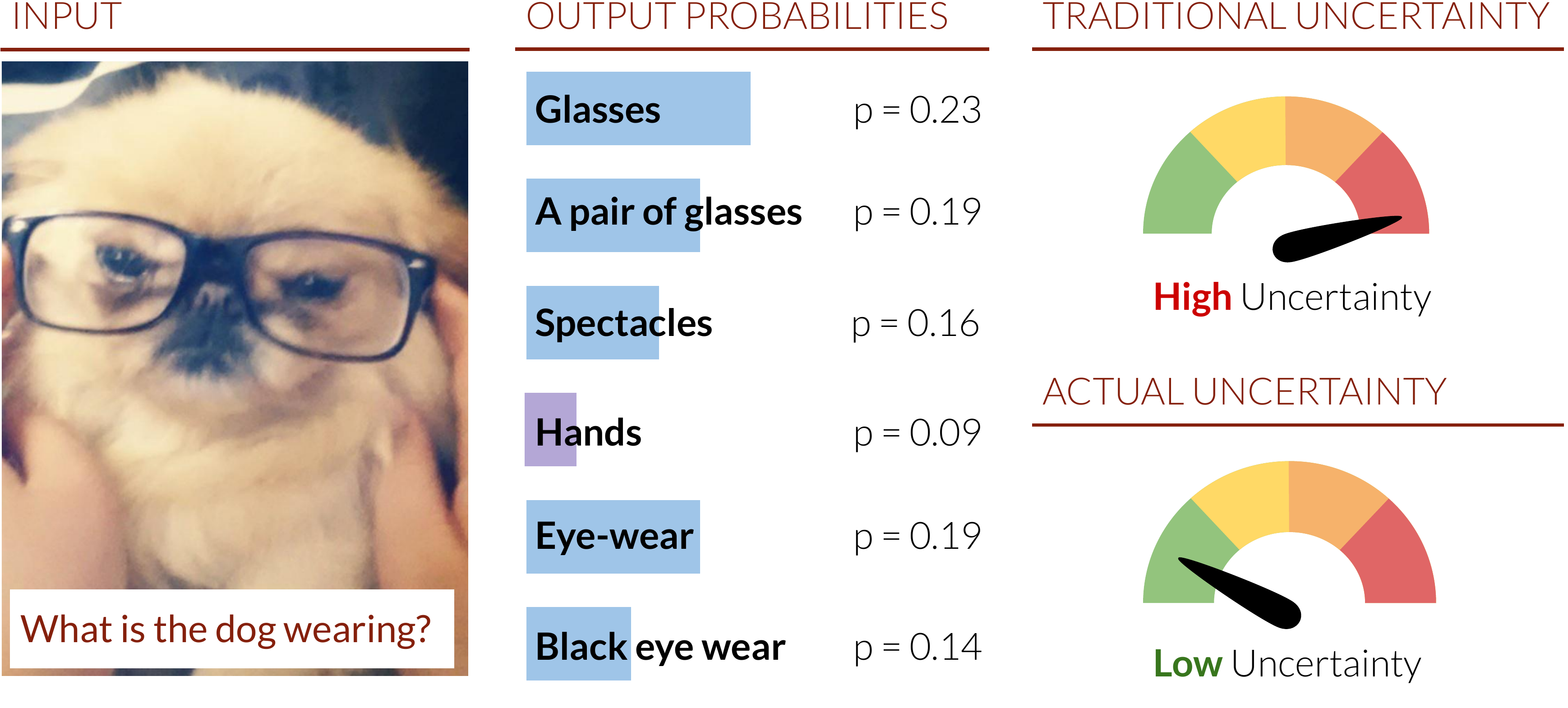}
\caption{Many vision tasks, such as image captioning and visual question answering, have multiple correct answers for the same input. In this example, multiple correct answers exist: ``Glasses'', ``A pair of Glasses'', etc. Existing active learning strategies overestimate uncertainty as they fail to account for paraphrases. We propose a new uncertainty estimation framework built over a visual-semantic embedding space that extends Monte-Carlo dropout-based Bayesian uncertainty for VQA and demonstrate that it is more sample efficient across multiple datasets.}
\label{fig:pull_figure}
\end{center}
\end{figure}

Active Learning is an approach for data annotation in supervised learning problems that selectively seeks labels for informative examples~\cite{tong2001support, settles2012active}. It has proven to be a potential solution~\cite{tong2001support} to maximize learning while reducing the data annotation cost for many problems, including vision-language generation tasks like Visual Question Answering (VQA)~\cite{misra2018learning, deng2018adversarial, lin2017active} or image captioning~\cite{shen2019learning} that rely on large, expensive, curated datasets~\cite{antol2015vqa,krishna2017visual,zhu2016visual7w,gurari2018vizwiz}. Nonetheless, applying existing active learning strategies to language generation tasks does not result in performance increase compared to randomly annotating examples~\cite{figueroa2012active,mairesse2010phrase,settles2008active}.

Traditionally designed for classification and regression~\cite{settles2012active}, popular active learning strategies use some notion of model output based uncertainty to select informative data~\cite{joshi2009multi,abramson2004active,collins2008towards,gal2017deep,kendall2017uncertainties}. There are two main challenges with existing approaches when they are adapted for vision-language tasks. First, they are founded on the assumption that no two output categories can be associated with any given input, assuming that the model's output space is a set of mutually exclusive labels. Unfortunately, this assumption is invalid for numerous vision-language tasks, such as image captioning and VQA, as one image or image-question pair can have multiple correct captions or answers~\cite{gurari2018vizwiz,bhattacharya2019does,yang2018visual}. This paraphrastic nature of language causes these measures to fail. Figure~\ref{fig:pull_figure} illustrates an instance of this phenomenon where the question ``what is the dog wearing?'' can be answered by ``Glasses'', ``Spectacles'', ``Eye-wear'' and countless other paraphrases. Measures like entropy~\cite{shen2017deep}, margin~\cite{culotta2005reducing}, or least confidence~\cite{shen2017deep} confound multiple correct answers to imply high uncertainty in the model's output.

Second, extending existing uncertainty functions for language generation models quickly renders them computationally intractable. Consider a VQA task with a model that can generate answers up to a length of $N$ with each word belonging to a vocabulary of size $\vert V \vert$. The output space is of the order of $N^{\vert V \vert}$. So, measuring uncertainty in model outputs grows exponentially with the vocabulary size. To circumvent intractability, Monte-Carlo simulations or similar approximations are often used to reduce the output space~\cite{settles2012active, Settles:2008:AAL:1613715.1613855}. However, there is no clear extension of such approximations for discrete tokens, like the words in an answer or a caption.

In this paper, we propose a novel active learning uncertainty sampling paradigm that estimates uncertainty in an embedding space instead of an output probability space. Our proposed method enables the use of dropout-based Bayesian Monte-Carlo uncertainty estimation~\cite{gal2016dropout} for sequence generation. We claim that our strategy is effective, even in the presence of multiple correct outputs with discrete words. We showcase the efficacy of our approach in a VQA setting, as it represents a natural manifestation of the phenomenon of having multiple correct candidates for a given input, but insist that our solution is extendable to any language generation task, provided that a semantic space can be created. Instead of immediately generating answers, we impose a semantic structure on the hidden representation of existing VQA models. Our semantic space is designed to embed paraphrases close together. It allows us to abstract away from the discrete tokens generated by VQA models and estimate uncertainty in the embedding space instead.


We show the utility of our uncertainty sampling strategy for VQA generation on the Visual Genome v1.4~\cite{krishna2017visual} and the VQA v2.0~\cite{antol2015vqa} datasets in an active learning setting. We observe that our solution achieves state-of-the-art results in multiple language generation metrics compared to existing uncertainty sampling strategies~\cite{shen2017deep, culotta2005reducing, scheffer2001active}, while being $5$ times more cost-efficient on Visual Genome and $3$ times more cost-efficient on VQA 2.0.

\section{Related Work}
We explore the field of active learning and comment on how existing uncertainty measurements fail in environments where we have multiple correct outputs. We place our work in context with the task of VQA, as we believe it represents a highly representative manifestation of these environments. Next, we dive into visual-semantic embedding, a practical solution to construct semantic spaces using visual priors. 
Finally, we explore Bayesian uncertainty measurement techniques.

\textbf{Active learning.} A typical active learning setting starts with an initial small set of labeled data $D_L$ and a large set of unlabeled data $D_U$. Knowing that labeling data is costly and time consuming, the task is to minimize the number of samples to annotate from $D_U$ while maximizing performance on a given task that requires those labels~\cite{tong2001support}. Active learning strategies have been successfully applied to a wide number of machine learning tasks, including image recognition~\cite{joshi2009multi, sener2017active}, information extraction~\cite{scheffer2001active,finn2003active,jones2003active,culotta2005reducing}, named entity recognition~\cite{shen2017deep,hachey2005investigating} and text categorization~\cite{lewis1994sequential, hoi2006batch}. Active learning strategies have included uncertainty-based sampling~\cite{joshi2009multi,abramson2004active,collins2008towards} and information gain~\cite{houlsby2011bayesian}. Others have introduced a theoretical dropout-based framework to measure uncertainty~\cite{gal2017deep,kendall2017uncertainties}. Similar solutions have been adapted to NLP tasks like Named Entity Recognition~\cite{shen2017deep} and Neural Semantic Parsing by adding Gaussian noise to the network weights~\cite{dong2018confidence}. In this paper, we empirically show how previous uncertainty sampling strategies do not perform better than random sampling in a VQA setting and propose a novel strategy that uses Bayesian uncertainty in a semantically structured embedding space.

\textbf{Visual question answering.} VQA systems expect an input image and a natural language question and attempt to output the correct answer~\cite{antol2015vqa}. VQA has received a considerable amount of attention in recent years with the development of several datasets, proposed as benchmarks~\cite{antol2015vqa, malinowski2015ask,johnson2017clevr,goyal2017making,krishna2017visual, ren2015exploring, zhu2016visual7w}, and of various models\cite{antol2015vqa, fukui2016multimodal, lu2016hierarchical, zhou2015simple, yang2016stacked, zhu2016visual7w, jabri2016revisiting, malinowski2014multi, wu2016ask}. To tackle the task, many proposed architectures encode and then merge the visual and textual information in order to classify answers~\cite{malinowski2015ask,ma2016learning, jabri2016revisiting}. Attention mechanisms have proved to be successful in this task~\cite{shih2016look, xu2016ask,lu2016hierarchical, anderson2018bottom}. Other work focuses on designing effective multi-modal feature fusion schemes~\cite{ben2017mutan,fukui2016multimodal,kim2016hadamard}.  While the performance has been encouraging, such approaches require a large amount of labelled data with a predefined, mutually exclusive set of answer categories. We explore a more realistic variant of VQA where models generate natural language answers, resulting in the generation of paraphrases.

\textbf{Semantically structured embeddings.} Our key insight lies in moving uncertainty estimation from the model's output space to a semantically structed internal embedding space. Since we specifically study vision-language tasks, we build a visual-semantic space that combines both visual and textual information in order to create a unified latent representation. This problem has been studied extensively in the last few years with work that involves jointly embedding images and text, at the word level~\cite{frome2013devise, kong2014you, joulin2016learning, matuszek2012joint, klein2015associating, socher2010connecting}, and sentence level~\cite{zitnick2013learning, karpathy2015, karpathy2014, Chen_2015_CVPR, kiros2014unifying, reed2016learning}. Visually grounded text representations have been applied to different tasks, including caption generation~\cite{kiros2014unifying, karpathy2015}, image retrieval~\cite{lin2014visual}, and visual question answering~\cite{malinowski2015ask}. Our approach is inspired by previous work~\cite{karpathy2014}, where we use image regions with their associated captions and adapt a margin objective to build the semantic space. We use the  semantic space to enable a new active learning strategy.

\textbf{Uncertainty.} As uncertainty estimation represents the most popular sampling strategy for active learning, we explore uncertainty estimation techniques developed for deep learning models. Even though Deep Learning systems are performant, they  remain uninterpretable, poor at quantifying predictive uncertainty, and overconfident in their predictions. 
To mitigate this problem, Bayesian models have been proposed by placing a prior distribution over model weights~\cite{williams1996gaussian, neal2012bayesian, mackay1992bayesian}. Yet, even though the problem is simple to formulate, deriving a posterior distribution for Deep Bayesian Neural Networks (BNN) is intractable which makes Bayesian inference difficult~\cite{gal2016dropout}. Therefore, focus has shifted to approximating BNNs with variational inference. Bayesian modeling of stochastic processes introduced new techniques into the field such as sampling-based and stochastic variational inference~\cite{blei2017variational,blundell2015weight}. 

On the other hand, Monte-Carlo Dropout~\cite{gal2017deep, gal2016dropout} has empirically demonstrated comparable uncertainty estimation quality to variational inference~\cite{gal2017deep}. A model trained with dropout can be used as a Bayesian model by making multiple predictions while sampling different dropout masks for each forward propagation. Estimating the posterior amounts to computing the mean and variance of the predictions. While previous work has showed promise in classification and regression tasks~\cite{yang2015multi, gal2016dropout, settles2012active}, such methods have not been extended to language generation. In fact, most language-based uncertainty work adapts classification-based uncertainty techniques to language tasks, transforming the problem from an open language problem to a constrained closed problem~\cite{shen2017deep,hachey2005investigating,lewis1994sequential,hoi2006batch,dong2018confidence}. We propose a new sampling strategy that extends Monte-Carlo dropout uncertainty~\cite{gal2016dropout} to measure uncertainty in a semantic space.

\section{Method}
We design an uncertainty measurement for active learning even when questions have multiple correct answers. In this section, we formulate our overall active learning framework, and then describe our uncertainty measurement approach. Our approach depends on two components: a semantic space that structures the model's output representations and a denoiser that refines our representations to get an accurate dropout-based Bayesian uncertainty estimation.

\subsection{Visual question answer generation}
We specifically study the scenario with multiple correct outputs using the VQA task, which expects an image-question pair $(q, i)$ as input and natural language answer $a$ as output. The goal is to train a model $\hat{a} \sim f(a| q,i)$, where $\hat{a}$ is the generated answer and $f(\cdot)$ is the trained VQA model. Since different sentences can have the same meaning, we define $S_a$, the set of all sentences semantically similar to $a$. When evaluating the performance of the VQA model, we consider any answer $\hat{a} \in S_a$ as a correct candidate for an answer. For example, in Figure~\ref{fig:pull_figure}, the question ``What is the dog wearing?'' can have multiple correct answer candidates, including ``Glasses'', ``Spectacles'', etc.

\begin{figure}[t]
\begin{center}
\includegraphics[width=\linewidth]{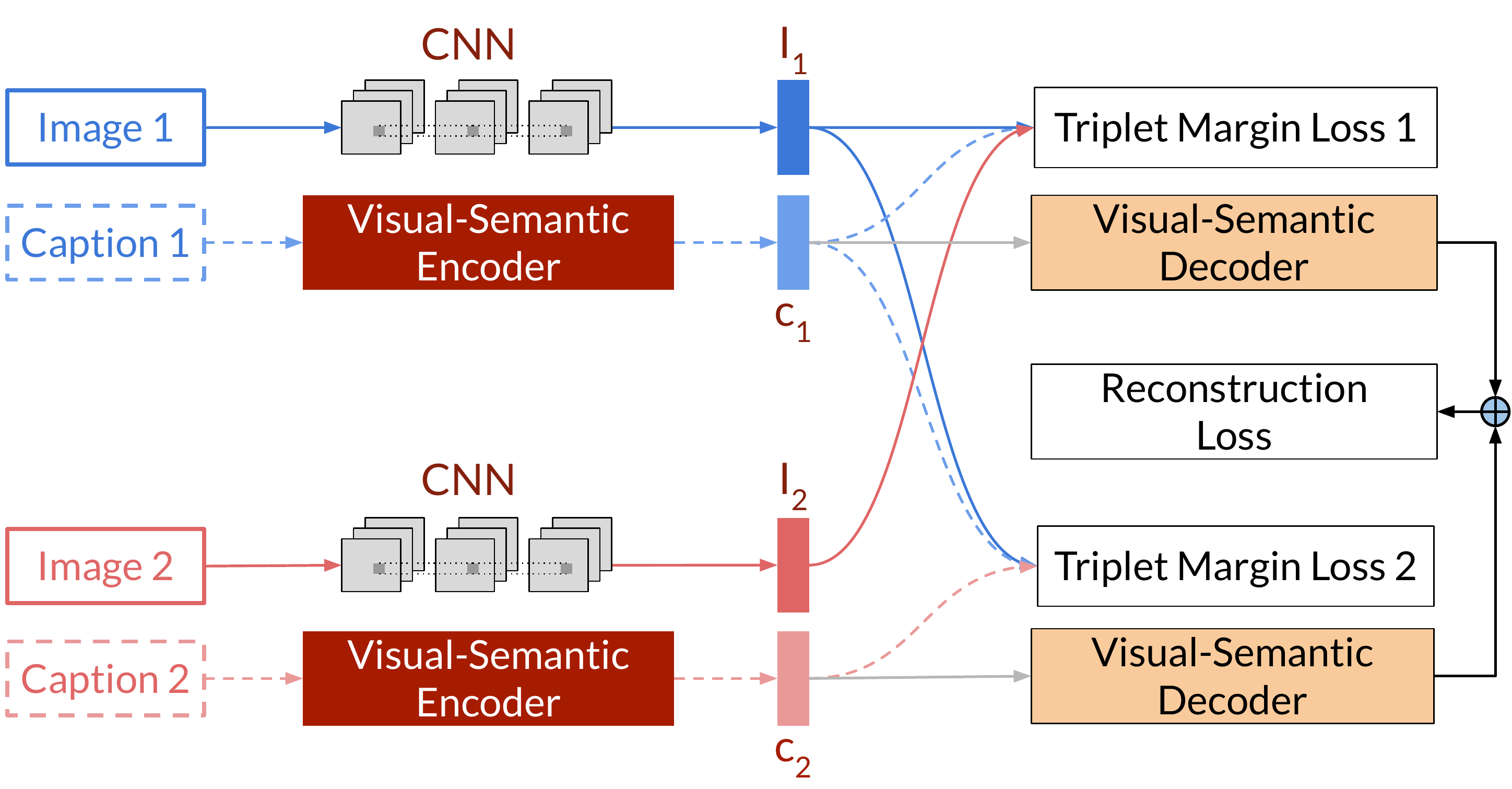}
\caption{We build a Visual-Semantic embedding using a small amount of image-caption pairs from Visual Genome. We optimize a contrastive loss such that similar language descriptions are mapped closer together via semantically similar image regions.}
\label{fig:vse_training}
\end{center}
\end{figure}

\begin{figure*}[t!]
    \begin{center}
    \includegraphics[width=\linewidth]{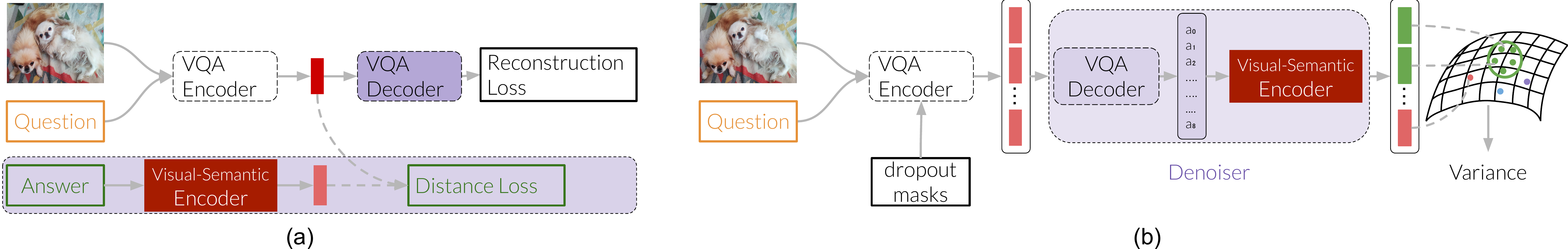}
    \label{fig:denoiser}
    \end{center}
\vspace{-2em}
\caption{(a) We train existing VQA models, which almost always follow an encoder-decoder architecture, with an additional embedding loss that structures the output according to the visual-semantic space. (b) We define an embedding denoiser, by combining the VQA decoder with the pretrained visual-semantic encoder. The denoiser refines these representations to get an accurate uncertainty estimation. }
    \label{fig:vqa}
\end{figure*}

Traditionally, VQA classifies the answer in a mutually exclusive set of $1000$ correct answers~\cite{antol2015vqa}. However, we study a more realistic variant of VQA where our model generates the answer in natural language. So, in all our experiments, we use the popular bottom-up attention model~\cite{anderson2018bottom} and replace its classifier with a long short term memory network (LSTM)~\cite{sundermeyer2012lstm} to generate answers instead of classifying them. Although we report results with the bottom-up attention model, our results are consistent with other existing VQA models~\cite{yang2016stacked}. Almost all VQA models follow a traditional encoder-decoder architecture, where $(q, i)$ are encoded to generate a hidden representation $z$. This representation is then decoded to produce the answer $\hat{a}$.

\subsection{Active learning for VQA}
We follow a traditional pool-based active learning setting~\cite{Settles:2008:AAL:1613715.1613855, settles2012active}. The model is initially trained with a small bootstrapping training set $\{(q, i, a)\} \in D_{train}$ to produce a starting model $f^{0}$. At every time step $t$, we receive a new pool $P^{t}$ of $N$ question-image pairs $(q, i)$ and must choose $K$ pairs to annotate using an oracle. The $K$ pairs are chosen using an uncertainty measurement $\{(q_1, i_1) \ldots (q_k, i_k)\} \sim U(P^t)$. Once the oracle answers the chosen questions, they are added to $D_{train}$ and the model is re-trained to $f^{t}$.

The process of incorporating new annotations by choosing data points from a pool continues for $T$ iterations. Our final model is $f^T$. In our experiments, we compare and evaluate different uncertainty measurements by the performance of their resulting $\{f^0, \ldots, f^T\}$ models. 

\subsection{Our contribution: Bayesian uncertainty as variance in semantic space}

As discussed earlier, traditional uncertainty estimation techniques for language generation are based on the assumption that all outputs must be mutually exclusive, and thus, fail when paraphrases exist. Also since the existing methods take exponentially long to compute with respect the size of the vocabulary~\cite{settles2012active}, these methods require approximations that lead to further errors~\cite{settles2012active}.

Our solution tackles both of these challenges. First, we build a visual-semantic space that captures semantic language similarly and maps similar paraphrases close together, and second, uses Monte-Carlo dropout-based Bayesian uncertainty~\cite{gal2016dropout} with a denoiser to measure the model's uncertainty. 

Monte-Carlo dropout~\cite{gal2016dropout} is an approximate inference approach to Bayesian Neural Networks that approximates a Gaussian Process by enforcing model dropout in both training and test time. At test time, due to the randomness induced by dropout, our model inference becomes stochastic which makes our output potentially variable. Applying a Monte-Carlo process with $m$ simulations, equivalent to applying $m$ forward passes each with a different dropout mask, results in a probabilistic distribution that can be used as a Bayesian interpretation of a neural network and allows the use of variance as an uncertainty measure.

\begin{figure*}[t]
\begin{center}
\includegraphics[width=\linewidth]{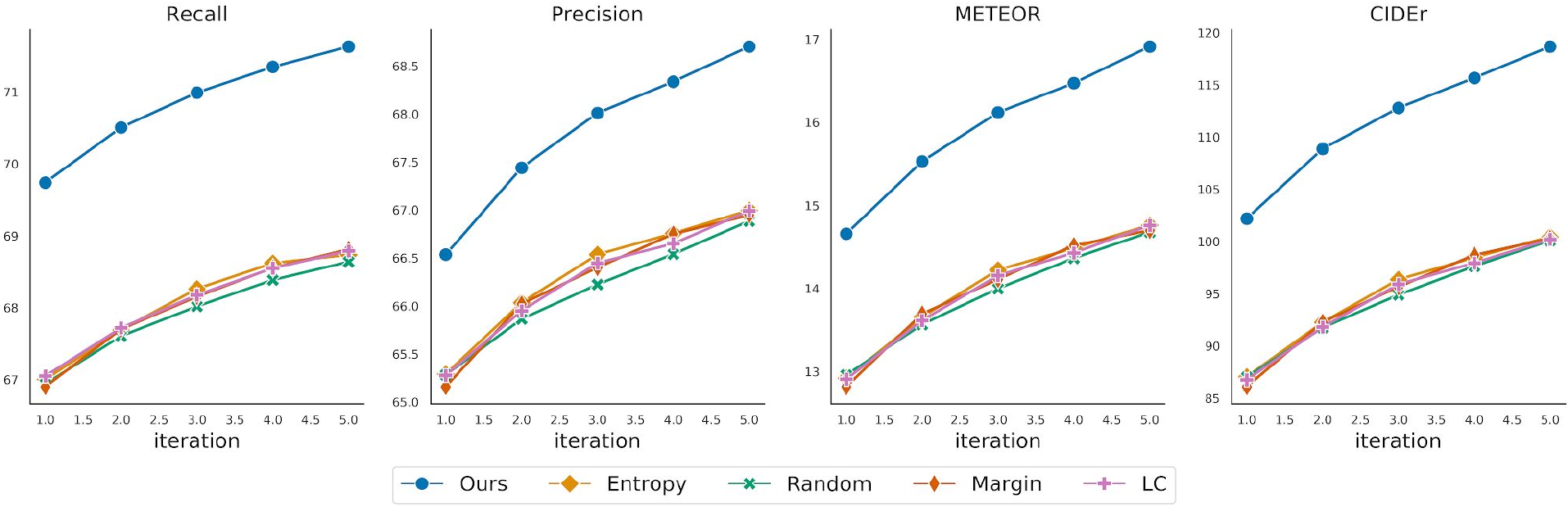}
\caption{Active learning strategies' performance on Visual Genome~\cite{krishna2017visual} measured using multiple language generation metrics. Performance of the initial model is not shown as it is the same for all strategies. We outperform all existing strategies. Additional metrics with exact numbers are included in the appendix.
}
\label{fig:alvg}
\end{center}
\end{figure*}

Instead of dealing with intractable output spaces, we measure uncertainty within a visual-semantic space that is computationally independent of the vocabulary size. More specifically, given an input $(q, i)$, we encode the pair into a hidden representation $h$ using $m$ forward passes, each with a randomly sampled dropout mask, resulting in $m$ hidden representations. We assume that the representations embed into a visual-semantic space, a space where similar visual and language concepts lie close together. We measure uncertainty as the sum of variances across all the dimensions of the representation vectors:
\begin{align}
    u = \textrm{Var}(\{h_1, h_2, \ldots, h_m\})
    \label{eq:1}
\end{align}
The $(q, i)$ inputs with the $K$ highest uncertainty scores are chosen and sent to an oracle for annotation. 

Intuitively, our measurement outputs a small uncertainty score when the $m$ forward passes all produce representations that occupy a small volume in the embedding space, implying that all the answers the model produces are paraphrases or at least semantically similar. Similarly, our measurement outputs a large uncertainty score when the $m$ representations occupy a large volume in the embedding space, implying that the answers produced by the model are different.

\subsection{Training the visual-semantic space}
Since we are demonstrating our method using a vision-language task, we build a visual-semantic space to structure language outputs. Similar to previous work, we design the embedding space using a dataset of image regions and their corresponding natural language captions~\cite{kiros2014unifying, frome2013devise,kong2014you,joulin2016learning,matuszek2012joint,klein2015associating,socher2010connecting,zitnick2013learning,karpathy2015,karpathy2014,Chen_2015_CVPR,reed2016learning}. Concretely, given a dataset of $(i, c)$ pairs of images and captions, we first embed the images using a pretrained ResNet50~\cite{he2016deep} $h_i = CNN(i)$ and the caption using a visual-semantic encoder $h_c = enc_{vs}(c)$. Next, we use contrastive loss, often called triplet loss, to ensure that similar image regions and similar captions are embedded nearby:
\begin{align}
    L_{image} &= [d(h_{c_j}, h_{i_j})- d(h_{c_j}, h_{i_k}) + \delta_m]_+\\
    L_{lang} &= [d(h_{i_j}, h_{c_j}) - d(h_{i_j}, h_{c_k}) +\delta_m]_+\\
    L_{recon} &= \mathcal{H}(dec_{vs}(h_{c_j}), c_j)\\
    L_{vs} &= \lambda_1 (L_{image} + L_{lang}) + \lambda_2 L_{recon}
\end{align}
where $(i_j c_j)$ and $(i_k, c_k)$ are two distinct image-captions pairs and $\delta_m$ is a hyper-parameter representing our loss margin. $d(\cdot, \cdot)$ is the euclidean distance function though any other distance function can also be used. $[\cdot]_+$ lower bounds all values to zero. To ensure that our embedding space does not collapse, we add a cross entropy loss $\mathcal{H}(\cdot)$ which reconstructs the caption from its embedding. $\lambda_1, \lambda_2$ are hyperparameters we optimize using a validation set of image-caption pairs. Figure~\ref{fig:vse_training} visualizes this training process.

\subsection{Training VQA using visual semantic space}
Finally, with the visual-semantic space trained, we can use it to train a VQA model and structure its outputs. Almost all VQA models proposed in Computer Vision follow an encoder-decoder architecture with an encoder which embeds the question-image pair $h \sim enc(q, i)$ and then a decoder that converts the embedding into an answer $a \sim dec(h)$~\cite{yang2016stacked,anderson2018bottom}. We train our model in a traditional way by imposing the following reconstruction cross entropy loss along with a visual-semantic embedding loss:
\begin{align}
    L_{vqa} &= L_{recon-vqa} +  L_{embed}\\
    L_{recon-vqa} &= \mathcal{H}(dec_{vqa}(enc(q, i)), a)\\
    L_{embed} &= \frac{\sum_{j = 1}^{n} (h_{a_j} - enc(q, i))^2}{n}
\end{align}
where $n$ is the size of the dataset and $h_{a_j} = enc_{vs}(a_j)$ is the projected answer embedding in the visual-semantic space. Figure~\ref{fig:vqa}(a) visualizing the training objective.

\subsection{Denoising the representations for uncertainty estimation}
Our approach depends on two central components: a semantic space that structures the model's output representations and a denoiser that refines our representations to get an accurate uncertainty estimation.
As discussed previously, we use the $L_{embed}$ distance loss with respect to a visual-semantic space to semantically structure our VQA encoder's hidden representation. In practice, the resulting representation is an approximation of the reference visual-semantic space. We find that embeddings for some concepts often result in a noisy representations. In our dropout-based framework, we find that variance measured on these noisy representations does not guarantee an accurate model uncertainty estimation and results in random sampling behavior.

\begin{figure*}[t]
\begin{center}
\includegraphics[width=\linewidth]{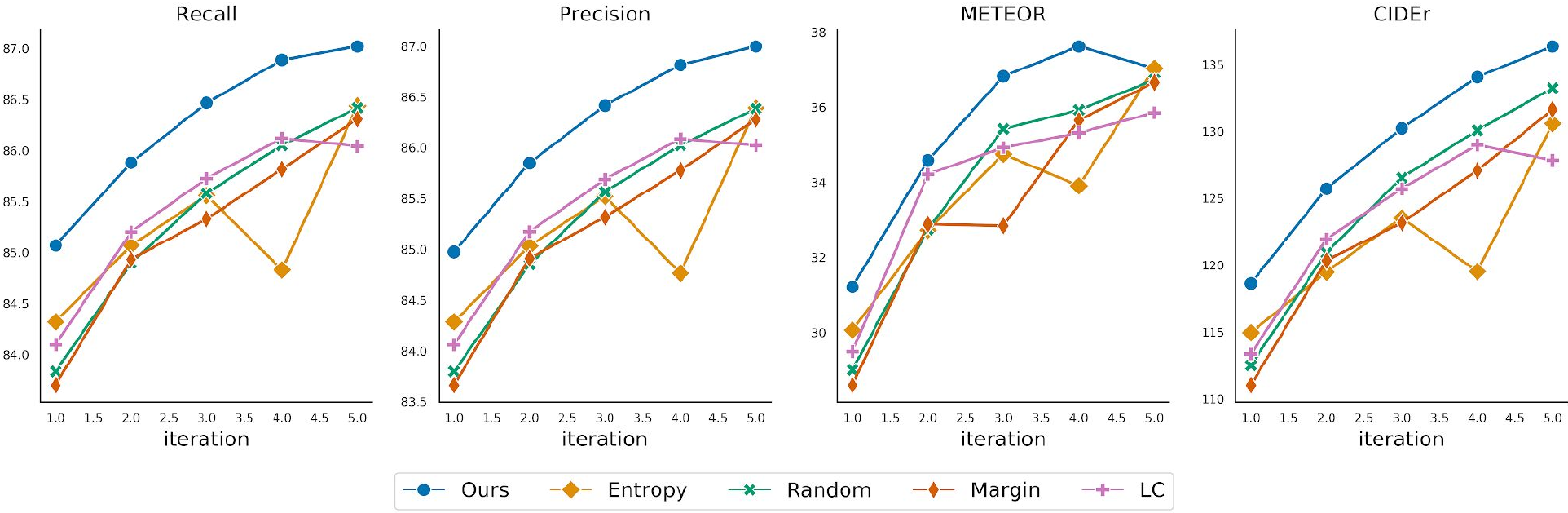}
\caption{Active learning strategies' performance on VQA~\cite{antol2015vqa} measured using multiple language generation metrics. Performance of the initial model is not shown as it is the same for all strategies. We outperform all existing strategies. Additional metrics with exact numbers are included in the appendix.}
\label{fig:alvqa}
\end{center}
\end{figure*}

To refine the image-question pair representations to be similar to the answer representation, we introduce an embedding denoiser module. The denoiser consists of components we have already introduced during training: the pretrained VQA decoder and the pretrained visual-semantic encoder. As the VQA decoder is trained to map the image-question representations to text, it learns the inherent noise in the embedding space generated by the VQA encoder. By decoding these concepts to answer outputs and then using the visual-semantic encoder to re-project them back, we semantically correct the embeddings. Figure~\ref{fig:vqa}(b) visualizes the denoiser. As the denoiser does not expect the VQA model's hidden representation to be semantically structured, we might expect that it renders the training VQA models with $L_{embed}$ loss unnecessary. However, we find in practice that the semantic information provided by $L_{embed}$ combined with the denoiser, not only increases the quality of the samples but also improves model performance (see Section~\ref{sec:ablations} for ablations). 

\subsection{Implementation Details}
We implement our visual-semantic model by combining a Resnet50~\cite{he2016deep} encoder for the image, with a bidirectional LSTM encoder for the question, with a hidden size of $512$ and $2$ layers. For our decoder, we use a 2-layer LSTM decoder with a hidden size of $512$. We train our the model using $70k$ image region and corresponding description pairs from Visual Genome~\cite{krishna2017visual}. We also feed VQA image-answer pairs from the VQA training set $10\%$ of all iterations ensure that answer concepts are mixed together with the captions. The model is optimized using Adam~\cite{kingma2014adam} with a learning rate of $1e^{-3}$ with zero weight decay. We initialize our LSTM embeddings using GLoVe~\cite{pennington2014glove} and train the model for a total of $27k$ iterations with a batch size of $128$.
When training our VQA model~\cite{anderson2018bottom}, we find that a dropout rate of $0.5$ results in good performance along with reasonable hidden representations for our uncertainty framework. To estimate our uncertainty, we simulate $m=20$ forward passes with dropout $0.5$ to generate our embedding distribution. If we increase $m$ beyond $20$, we find that the change in the uncertainty value is negligible.

\section{Experiments}

In our experiments, we empirically demonstrate that Bayesian uncertainty, measured as variance in the visual-semantic space, is a better sampling strategy for active learning than existing strategies. Furthermore, we show that all existing uncertainty measures perform better when the output of the VQA model is structured with a visual-semantic space.

\noindent\textbf{Datasets.} We study our approach using two existing VQA datasets: Visual Genome~\cite{krishna2017visual} and VQA~\cite{antol2015vqa}.
Visual Genome contains $108$k images densely annotated with scene graphs containing objects, attributes and relationships, as well as region descriptions and visual question answers. We use $70$k of $5$M randomly sampled region descriptions to create our visual-semantic space. For the VQA model, we use the $1.18$M question answers. We split our data into a train set of $69k$ images with $853$k question answers and a validation set of $27$k images with $335$k $(q,a)$ pairs. VQA is a dataset of $123$k MSCOCO~\cite{lin2014microsoft} images with $658$k visual question answers. We use the default train/val split with a training set of $83$k images with  $443$k question answers and a test set of $40$k images with $215$k $(q,a)$ pairs.

\noindent\textbf{Active learning setup.} In order to showcase the advantage of usin
g our uncertainty measurement, we test its efficiency against widely used uncertainty sampling strategies~\cite{culotta2005reducing, scheffer2001active, shen2017deep}. We setup our active learning pipeline as follows: we randomly initialize our bootstrapping set $D_{train}$ with size representing $5\%$ of the original training set, and pretrain our model $f^0$. $f^0$ is then used as the starting point for all our experiments. For each active learning iteration, we sample a pool $P^t$ with size representing $15\%$ of the original dataset size. Using our sampling strategies, we update $D_{train}$ with the $K$ best scoring data points such that we have an increase of $5\%$ of the entire dataset in each iteration. We retrain our model using the resulting train set and repeat the procedure for $5$ iterations, resulting in a final $D_{train}$ size of $30\%$ of the original dataset. We train the visual-semantic space with the same amount of image-caption pairs as the question-answer pairs the initial $f^0$ uses.

\begin{table}[t!]
	\caption{We report CIDEr performance per question type for our measure, as well as \texttt{Margin} sampling. We also report corresponding samples distributions.}
	\label{tab:categ}
    \begin{center}
        \small
		\setlength{\tabcolsep}{.10em}
		\scalebox{0.85}{
		\begin{tabular}{lrlcccccc}
			\hline
			& & & \multicolumn{5}{c}{Active learning iteration} \\
			\cline{4-8}
			&Q-Type & Model & 1 &2  &3 &4 &5\\
			\hline
			\multirow{10}{*}{\rotatebox[origin=l]{90}{Performance}}
            & \multirow{2}{*}{What} & Ours &\textbf{97.60} & \textbf{104.38} &  \textbf{110.17} &  \textbf{112.47} & \textbf{114.62}\\&& Margin& 84.98 & 86.47 & 95.00 &  98.75 & 99.44\\\cline{3-8}
		    &\multirow{2}{*}{Where} & Ours & \textbf{100.77} & \textbf{107.06} & \textbf{112.73} & \textbf{114.35} & \textbf{116.62} & \\ && Margin & 85.57 & 92.01 & 95.09 & 99.56 &  100.48 \\\cline{3-8}
			&\multirow{2}{*}{Who} & Ours & \textbf{104.79}  & \textbf{111.27}  & \textbf{116.92}  & \textbf{119.31} & \textbf{121.2}\\&& Margin & 84.98 &  92.07 & 96.84 &  99.21 & 98.92\\\cline{3-8}
            &\multirow{2}{*}{How} & Ours & \textbf{98.14} & \textbf{104.84} & \textbf{111.23} &  \textbf{112.45} & \textbf{114.77} & \\ && Margin & 84.40 &  90.28 &  95.79 &  100.36 & 97.08\\\cline{3-8}
			&\multirow{2}{*}{How many} & Ours & \textbf{104.95} & \textbf{110.60} & \textbf{115.98} & \textbf{117.13} & \textbf{120} & \\ && Margin & 86.15 &  91.44 &  93.55 &  99.34 & 101.87\\
			\hline
			\multirow{10}{*}{\rotatebox[origin=c]{90}{Sampling \%}}
            &\multirow{2}{*}{What} & Ours & \textbf{62.80\%} & \textbf{60.50\%} & \textbf{70.00\%} & \textbf{60.30\%} & \textbf{61.60\%} &\\&& Margin & 53.20\% & 50.10\% & 49.20\% & 48.30\% & 49.50\%\\\cline{3-8}
		    &\multirow{2}{*}{Where} & Ours & 20.30\% & 21.60\% & 22.20\% & 21.40\% & 22.70\% & \\ && Margin & \textbf{25.90\%} & \textbf{28.80\%} & \textbf{29.00\%} & \textbf{28.50\%} & \textbf{28.40\%}\\\cline{3-8}
			&\multirow{2}{*}{Who} & Ours & 3.60\% & 4.80\% & 3.80\% & 4.80\% & 4.00\% & \\ && Margin & \textbf{5.30\%} & \textbf{5.90\%} & \textbf{6.10\%} & \textbf{6.90\%} & \textbf{6.90\%}\\\cline{3-8}
            &\multirow{2}{*}{How} & Ours & \textbf{6.10\%} & \textbf{6.30\%} & \textbf{5.40\%} & \textbf{6.50\%} & \textbf{5.20\%} & \\ && Margin & 4.80\% & 4.50\% & 4.50\% & 4.20\% & 4.10\%\\\cline{3-8}
			&\multirow{2}{*}{How many} & Ours & 1.40\% & 1.00\% & 0.90\% & 1.00\% & 1.00\% & \\&& Margin & \textbf{5.30\%} & \textbf{5.40\%} & \textbf{6.20\%} & \textbf{7.20\%} & \textbf{6.60\%}\\
			\hline
		\end{tabular}
		}
	\end{center}
\end{table}

\begin{table*}[t!]
    \caption{We report our strategy's performance and find that it outperforms all baselines. We also notice that the baselines improve considerably if we add visual-semantic information (+\texttt{VS}). (+\texttt{Deno} refers to the denoiser.}
	\label{tab:server}
    \begin{center}\small
		\setlength{\tabcolsep}{.25em}
		\scalebox{0.77}{
		\begin{tabular}{l|ccccc|ccccc|ccccc|ccccc}
			\hline
			 & \multicolumn{5}{c}{Bert Recall} &  \multicolumn{5}{c}{Bert Precision} &  \multicolumn{5}{c}{METEOR} &
			 \multicolumn{5}{c}{CIDEr} \\
			\hline
			Iteration &  1 & 2 & 3 & 4 & 5 &    1 & 2 & 3 & 4 & 5 &     1 & 2 & 3 & 4 & 5 &     1 & 2 & 3 & 4 & 5\\
		\hline
			Dataset Size (\%)&  10 & 15 & 20 & 25 & 30 &    10 & 15 & 20 & 25 & 30 &     10 & 15 & 20 & 25 & 30 &     10 & 15 & 20 & 25 & 30\\ 
			\hline
			\hline
			\texttt{Random}&  66.94 & 67.61 & 68.02 & 68.39 & 68.64 &  65.28 & 65.86 & 66.22 & 66.54 & 66.89 &  12.96 & 13.57 & 14.00 & 14.37 & 14.68 &  87.01 & 91.75 & 94.90 & 97.63 & 100.09\\
			\texttt{Margin} &  66.90 & 67.70 & 68.16 & 68.55 & 68.81 &  65.16 & 66.02 & 66.4 & 66.75 & 66.95 &  12.82 & 13.70 & 14.11 & 14.52 & 14.70 &  86.10 & 92.28 & 95.66 & 98.70 & 100.30\\
			\texttt{LC} & 67.05 & 67.72 & 68.18 & 68.55 & 68.79 &  65.28 & 65.95 & 66.45 & 66.65 & 66.99 &  12.91 & 13.62 & 14.16 & 14.43 & 14.76 &  86.76 & 91.80 & 95.91 & 97.94 & 100.22\\
			\texttt{Entropy}& 67.00 & 67.69 & 68.26 & 68.62 & 68.74 & 65.29 & 66.04 & 66.54 & 66.76 & 67.00 & 12.92 & 13.66 & 14.22 & 14.49 & 14.76 & 87.05 & 92.26 & 96.36 & 98.47 & 100.45\\
			\hline
			\texttt{Random} + \texttt{VS} & 69.57 & 70.35 & 70.72 & 71.10 & 71.35 &  66.39 & 67.24 & 67.81 & 68.09 & 68.41 &  14.52 & 15.34 & 15.91 & 16.23 & 16.55 &  101.68 & 107.76 & 111.88 & 114.27 & 116.72 \\ 
			\texttt{Margin} + \texttt{VS} & 69.56 & 70.36 & 70.75 & 71.08 & 71.42 &  66.29 & 67.24 & 67.64 & 68.11 & 68.49 &  14.43 & 15.40 & 15.86 & 16.30 & 16.70 &  100.73 & 108.20 & 111.61 & 115.14 & 117.97\\
			\texttt{LC} + \texttt{VS} & 69.53 & 70.26 & 70.56 & 71.00 & 71.27 &  66.25 & 67.13 & 67.42 & 67.96 & 68.34 &  14.39 & 15.28 & 15.68 & 16.18 & 16.56 &  100.29 & 107.00 & 110.44 & 113.68 & 116.56 \\
			\texttt{Entropy} + \texttt{VS} &69.47 & 70.31 & 70.68 & 71.07 & 71.37 &  66.17 & 67.22 & 67.67 & 68.10 & 68.38 &  14.31 & 15.33 & 15.83 & 16.27 & 16.56 &  100.26 & 108.02 & 111.49 & 114.66 & 116.98\\
			\hline
			\texttt{Ours} (\texttt{Baye})& 67.15 & 67.76 & 68.14 & 68.46 & 68.73 & 65.47 & 65.99 & 66.35 & 66.70 & 67.05 & 13.04 & 13.67 & 14.03 & 14.42 & 14.75 & 87.92 & 92.24 & 95.33 & 97.71 & 100.15 \\ 
			\texttt{Ours} (\texttt{Baye} + \texttt{Deno})& 67.13 & 67.86 & 68.38 & 68.68 & 68.98 & 65.49 & 66.17 & 66.63 & 66.93 & 67.22 & 13.05 & 13.83 & 14.31 & 14.67 & 14.96 & 87.59 & 93.01 & 96.81 & 99.34 & 101.44\\
			\texttt{Ours} (\texttt{Baye} + \texttt{VS})& 69.48 & 70.30 & 70.59 & 70.88 & 71.27 & 66.34 & 67.27 & 67.62 & 68.01 & 68.47 &  14.53 & 15.38 & 15.74 & 16.18 & 16.59 & 101.47 & 107.97 & 110.73 & 113.91 & 117.06\\
			\texttt{Ours} (\texttt{Baye} + \texttt{VS} + \texttt{Deno}) & \textbf{69.74} & \textbf{70.51} & \textbf{70.99} & \textbf{71.35} & \textbf{71.63} & \textbf{66.54} & \textbf{67.44} & \textbf{68.01} & \textbf{68.34} & \textbf{68.70}& \textbf{14.66} & \textbf{15.53} & \textbf{16.12} & \textbf{16.47} & \textbf{16.91} & \textbf{102.19} & \textbf{108.87} & \textbf{112.80} & \textbf{115.67} & \textbf{118.67}\\
			\hline
		\end{tabular}
		}
	\end{center}
\end{table*}

\noindent\textbf{Baselines.}
We compare our uncertainty measure against the most popular uncertainty-based strategies. All baselines use a traditional encoder-decoder architecture without imposing a visual-semantic embedding loss.
Random sampling (\texttt{Random}) is a passive learning strategy that queries points following a uniformly random distribution. 
Least confidence (\texttt{LC})~\cite{culotta2005reducing} queries the instances for which the model has the least confidence in its most likely generated sequence:
$u(q, i) = 1 - P(a^*| q, i),$
where $P(a^*| q, i)$ is the probability of the most confident model response. 
Margin Sampling (\texttt{Margin})~\cite{scheffer2001active}:  queries the instances with the least margin between the output probabilities for the two most likely generated answers of our model:
$u(q, i) = 1 + P(a_2| q, i) - P(a_1 | q, i)$,
where $P(a_j| Q, I)$ is the probability of the $j^{th}$ most confident model response, $j \in \{1, 2\}$. 
Maximum Entropy Sampling (\texttt{Entropy})~\cite{shen2017deep} queries the instances that maximize the entropy of our models output:
$u(q, i) = - \sum_{j=1}^{b} P_j(a|q, i) logP_j(a|q, i)$.
We use a beam size of $b=5$ to get our uncertainty score.

While other active learning strategies like information gain and density based methods exist, some require an explicit enumeration over the output space, which is intractable, while others are not scalable for large datasets and adapting them in language generation models is still an open research problem~\cite{settles2012active}. 

\noindent\textbf{Evaluation.} To evaluate answer quality, we use the standard automatic evaluation metrics, namely CIDEr~\cite{vedantam2015cider}, METEOR~\cite{denkowski2014meteor}, and BERT (precision and recall)~\cite{zhang2019bertscore}. We report additional metrics, like Bert F1~\cite{zhang2019bertscore}, Bleu~\cite{papineni2002bleu}, ROUGE-L~\cite{lin2004rouge} and accuracy in the appendix.

\subsection{Active learning performance}
\noindent\textbf{Setup.} We report quantitative results from our active learning experiments in Table~\ref{fig:alvg}, as well as in Figure~\ref{fig:alvqa}. To ensure that our results are not a product of noise, we run all experiments $5$ times for each strategy on both datasets. Each run is accompanied with a different random seed to alter how the weights of the model are initialized and which pool of image-question pairs arrives at every step. We present the average scores from all $5$ separate experiments.

\noindent\textbf{Results.} Our solution out-performs all traditional active learning strategies in all metrics by a large margin for both Visual Genome (see Table~\ref{fig:alvg}) and VQA 2.0 (see Table~\ref{fig:alvqa}). The increase in performance is larger for Visual Genome, which has more unique answers than VQA 2.0, resulting in the model learning more sequences to express the same answer. On Visual Genome, our measure is over $5$ times more cost efficient than most existing sampling strategies, performing at $20\%$ of the annotation budget better than what other strategies manage after using the whole budget across all metrics. 
Similarly for VQA 2.0, we find that our solution is around $3$ times more cost efficient, performing at $60\%$ of the annotation budget better than what other strategies manage after using the whole budget across all metrics. This demonstrates our framework's ability to maximize performance when multiple correct answers are present.

\subsection{What questions are sampled?}
\noindent\textbf{Setup.} Next, we dive into what types of questions are sampled by \texttt{Ours} versus \texttt{Margin}, which is the best performing baseline. We report the sampling statistics per question category, i.e~the percentage of a specific type of question the strategy chooses at every step. We also report the performance of the models on the test set for those specific categories. We only report CIDEr scores as we see the same trend across the other metrics. Results are reported in Table~\ref{tab:categ}. We look at ``what'', ``where'', ``who'', ``how'' and ``how many (counting)'' questions in particular as they make up more than $90\%$ of questions in Visual Genome. We also report how well the models perform at every iteration on the test split containing only that type of questions.

\noindent\textbf{Results.} Not only do we outperform \texttt{Margin} in cases where we sample more points for a specific type of question but also in cases where we sample less. In order to explain this behavior, we study samples from the $3^{rd}$ iteration of active learning, in which we sample significantly less ``where'' questions and more ``what'' questions and yet perform better than \texttt{Margin} on both types. We find that \texttt{Margin} samples $25.3\%$ more answers of length $\geq 4$ on ``where'' questions. We also look at the average length of answers in Visual Genome~\cite{krishna2017visual} and find that ``what'' questions have an average length of $1.9(\pm1.2)$ while ``where'' questions have an average length of $3.1(\pm1.5)$.  Longer answers typically have a higher potential of having paraphrases. We find this assumption to be true qualitatively: \texttt{Margin} samples questions with ground truth answers like ``next to the large brick buildings'', ``near the tall buildings'' and ``by the buildings'' when it already has $6$ paraphrases already in its training set. Furthermore, \texttt{Ours} samples $27\%$ more new concepts on ``what'' question that might be less prone to redundancy since they are generally shorter answers (refer to Appendix for more details).
We also see that our method samples very few counting questions as the model picks up on the dataset bias where the answer is usually ``2'' and therefore has a lower uncertainty. 

\subsection{Ablations}
\label{sec:ablations}
\noindent\textbf{Setup.} Our solution involves three components: a visual-semantic space (\texttt{VS}), applying dropout-based Bayesian uncertainty within that space (\texttt{Baye}), and utilizing a denoiser to make the measurements more accurate (\texttt{Deno}). The visual-semantic space, however, can be used during training for any VQA model to structure the outputs and doesn't need to specifically rely on dropout-based Bayesian uncertainty. So, we can add $L_{embed}$ as a loss and still utilize existing uncertainty functions. Here, we perform experiments by ablating the three components of our model as well as adding the visual-semantic space loss and reporting how they impact existing methods.

\noindent\textbf{Results.} We report the impact of the structure provided by the visual-semantic space when still using existing uncertainty measure in Table~\ref{tab:server}. We notice that even though our proposed strategy outperforms the extended traditional strategies, all existing uncertainty methods achieve a considerable boost in performance over iterations. This leads to two important conclusions: (1) structuring the output space of problems that have multiple correct answers can improve uncertainty estimations by all existing measures and (2) measuring uncertainty in the embedding space instead of model outputs proves to be more robust with multiple correct answers. Our ablations also demonstrate the importance of the denoiser and visual-semantic space when using Bayesian uncertainty estimation. We show that \texttt{Baye} by itself performs just as well as the existing baselines but combined with \texttt{Deno} or \texttt{VS} increases its performance and together \texttt{Baye+VS+Deno} performs the best.

\section{Discussion}
While promising, it is important to note that our method does have a few limitations. 
First, constructing a semantic space requires additional data. In our case, the availability of Visual Genome's~\cite{krishna2017visual} region descriptions made constructing a visual-semantic space feasible. However, in different tasks where such data might not be already available, such an approach can increase the overall cost. 
Second, it is possible for semantic embeddings for some concepts to occupy a larger convex hull than others. Therefore, variance as a measure might overestimate uncertainty for concepts that cover a larger convex hull while underestimating uncertainty for those placed within a smaller hull. Measuring uncertainty with respect to the density of specific concepts is an open research question we leave to future work.
Third, as we measure the variance on $m$ sampled embeddings as our uncertainty, our solution needs multiple forward passes. This gives us a runtime and computational disadvantage over existing baselines.
Even though these limitations exist, we consider them implementation limitations, left for further research work, and emphasize the higher level contribution of our paper which is the proposal of a new paradigm that investigates uncertainty as variance in semantic space.

\section{Conclusion}

Existing uncertainty sampling strategies in active learning are no better than random sampling for sequence generation tasks with multiple correct answers. We propose a novel uncertainty sampling paradigm that moves away from a probabilistic to an embedding-based uncertainty estimation and overcomes the paraphrastic nature of language. We evaluate our solution in an active learning setting for the VQA generation task and outperform existing sampling strategies. Our model samples fewer paraphrases and more novel concepts and is $5\times$ more cost-efficient on Visual Genome and $3\times$ on VQA 2.0.


\noindent\textbf{Acknowledgements.} This work was partially funded by the Brown Institute of Media Innovation and by Toyota Research Institute (``TRI'') but solely reflects the opinions and conclusions of its authors and not TRI or any Toyota entity.

{\small
\bibliographystyle{ieee}
\bibliography{references}

\begin{thebibliography}{10}\itemsep=-1pt

\bibitem{abramson2004active}
Y.~Abramson and Y.~Freund.
\newblock Active learning for visual object recognition.
\newblock In {\em Technical report}. UCSD, 2004.

\bibitem{anderson2018bottom}
P.~Anderson, X.~He, C.~Buehler, D.~Teney, M.~Johnson, S.~Gould, and L.~Zhang.
\newblock Bottom-up and top-down attention for image captioning and visual
  question answering.
\newblock In {\em Proceedings of the IEEE Conference on Computer Vision and
  Pattern Recognition}, pages 6077--6086, 2018.

\bibitem{antol2015vqa}
S.~Antol, A.~Agrawal, J.~Lu, M.~Mitchell, D.~Batra, C.~Lawrence~Zitnick, and
  D.~Parikh.
\newblock Vqa: Visual question answering.
\newblock In {\em Proceedings of the IEEE international conference on computer
  vision}, pages 2425--2433, 2015.

\bibitem{ben2017mutan}
H.~Ben-Younes, R.~Cadene, M.~Cord, and N.~Thome.
\newblock Mutan: Multimodal tucker fusion for visual question answering.
\newblock In {\em Proceedings of the IEEE international conference on computer
  vision}, pages 2612--2620, 2017.

\bibitem{bhattacharya2019does}
N.~Bhattacharya, Q.~Li, and D.~Gurari.
\newblock Why does a visual question have different answers?
\newblock {\em arXiv preprint arXiv:1908.04342}, 2019.

\bibitem{blei2017variational}
D.~M. Blei, A.~Kucukelbir, and J.~D. McAuliffe.
\newblock Variational inference: A review for statisticians.
\newblock {\em Journal of the American Statistical Association},
  112(518):859--877, 2017.

\bibitem{blundell2015weight}
C.~Blundell, J.~Cornebise, K.~Kavukcuoglu, and D.~Wierstra.
\newblock Weight uncertainty in neural networks.
\newblock {\em arXiv preprint arXiv:1505.05424}, 2015.

\bibitem{Chen_2015_CVPR}
X.~Chen and C.~Lawrence~Zitnick.
\newblock Mind's eye: A recurrent visual representation for image caption
  generation.
\newblock In {\em The IEEE Conference on Computer Vision and Pattern
  Recognition (CVPR)}, June 2015.

\bibitem{collins2008towards}
B.~Collins, J.~Deng, K.~Li, and L.~Fei-Fei.
\newblock Towards scalable dataset construction: An active learning approach.
\newblock In {\em European conference on computer vision}, pages 86--98.
  Springer, 2008.

\bibitem{culotta2005reducing}
A.~Culotta and A.~McCallum.
\newblock Reducing labeling effort for structured prediction tasks.
\newblock In {\em AAAI}, volume~5, pages 746--751, 2005.

\bibitem{deng2018adversarial}
Y.~Deng, K.~Chen, Y.~Shen, and H.~Jin.
\newblock Adversarial active learning for sequences labeling and generation.
\newblock In {\em IJCAI}, pages 4012--4018, 2018.

\bibitem{denkowski2014meteor}
M.~Denkowski and A.~Lavie.
\newblock Meteor universal: Language specific translation evaluation for any
  target language.
\newblock In {\em Proceedings of the ninth workshop on statistical machine
  translation}, pages 376--380, 2014.

\bibitem{dong2018confidence}
L.~Dong, C.~Quirk, and M.~Lapata.
\newblock Confidence modeling for neural semantic parsing.
\newblock {\em arXiv preprint arXiv:1805.04604}, 2018.

\bibitem{figueroa2012active}
R.~L. Figueroa, Q.~Zeng-Treitler, L.~H. Ngo, S.~Goryachev, and E.~P. Wiechmann.
\newblock Active learning for clinical text classification: is it better than
  random sampling?
\newblock {\em Journal of the American Medical Informatics Association},
  19(5):809--816, 2012.

\bibitem{finn2003active}
A.~Finn and N.~Kushmerick.
\newblock Active learning selection strategies for information extraction.
\newblock In {\em Proceedings of the International Workshop on Adaptive Text
  Extraction and Mining (ATEM-03)}, pages 18--25, 2003.

\bibitem{frome2013devise}
A.~Frome, G.~S. Corrado, J.~Shlens, S.~Bengio, J.~Dean, T.~Mikolov, et~al.
\newblock Devise: A deep visual-semantic embedding model.
\newblock In {\em Advances in neural information processing systems}, pages
  2121--2129, 2013.

\bibitem{fukui2016multimodal}
A.~Fukui, D.~H. Park, D.~Yang, A.~Rohrbach, T.~Darrell, and M.~Rohrbach.
\newblock Multimodal compact bilinear pooling for visual question answering and
  visual grounding.
\newblock {\em arXiv preprint arXiv:1606.01847}, 2016.

\bibitem{gal2016dropout}
Y.~Gal and Z.~Ghahramani.
\newblock Dropout as a bayesian approximation: Representing model uncertainty
  in deep learning.
\newblock In {\em international conference on machine learning}, pages
  1050--1059, 2016.

\bibitem{gal2017deep}
Y.~Gal, R.~Islam, and Z.~Ghahramani.
\newblock Deep bayesian active learning with image data.
\newblock In {\em Proceedings of the 34th International Conference on Machine
  Learning-Volume 70}, pages 1183--1192. JMLR. org, 2017.

\bibitem{goyal2017making}
Y.~Goyal, T.~Khot, D.~Summers-Stay, D.~Batra, and D.~Parikh.
\newblock Making the v in vqa matter: Elevating the role of image understanding
  in visual question answering.
\newblock In {\em Proceedings of the IEEE Conference on Computer Vision and
  Pattern Recognition}, pages 6904--6913, 2017.

\bibitem{gurari2018vizwiz}
D.~Gurari, Q.~Li, A.~J. Stangl, A.~Guo, C.~Lin, K.~Grauman, J.~Luo, and J.~P.
  Bigham.
\newblock Vizwiz grand challenge: Answering visual questions from blind people.
\newblock In {\em Proceedings of the IEEE Conference on Computer Vision and
  Pattern Recognition}, pages 3608--3617, 2018.

\bibitem{hachey2005investigating}
B.~Hachey, B.~Alex, and M.~Becker.
\newblock Investigating the effects of selective sampling on the annotation
  task.
\newblock In {\em Proceedings of the Ninth Conference on Computational Natural
  Language Learning}, pages 144--151. Association for Computational
  Linguistics, 2005.

\bibitem{he2016deep}
K.~He, X.~Zhang, S.~Ren, and J.~Sun.
\newblock Deep residual learning for image recognition.
\newblock In {\em Proceedings of the IEEE conference on computer vision and
  pattern recognition}, pages 770--778, 2016.

\bibitem{hoi2006batch}
S.~C. Hoi, R.~Jin, J.~Zhu, and M.~R. Lyu.
\newblock Batch mode active learning and its application to medical image
  classification.
\newblock In {\em Proceedings of the 23rd international conference on Machine
  learning}, pages 417--424. ACM, 2006.

\bibitem{houlsby2011bayesian}
N.~Houlsby, F.~Husz{\'a}r, Z.~Ghahramani, and M.~Lengyel.
\newblock Bayesian active learning for classification and preference learning.
\newblock {\em arXiv preprint arXiv:1112.5745}, 2011.

\bibitem{jabri2016revisiting}
A.~Jabri, A.~Joulin, and L.~Van Der~Maaten.
\newblock Revisiting visual question answering baselines.
\newblock In {\em European conference on computer vision}, pages 727--739.
  Springer, 2016.

\bibitem{johnson2017clevr}
J.~Johnson, B.~Hariharan, L.~van~der Maaten, L.~Fei-Fei, C.~Lawrence~Zitnick,
  and R.~Girshick.
\newblock Clevr: A diagnostic dataset for compositional language and elementary
  visual reasoning.
\newblock In {\em Proceedings of the IEEE Conference on Computer Vision and
  Pattern Recognition}, pages 2901--2910, 2017.

\bibitem{jones2003active}
R.~Jones, R.~Ghani, T.~Mitchell, and E.~Riloff.
\newblock Active learning for information extraction with multiple view feature
  sets.
\newblock {\em Proc. of Adaptive Text Extraction and Mining, EMCL/PKDD-03,
  Cavtat-Dubrovnik, Croatia}, pages 26--34, 2003.

\bibitem{joshi2009multi}
A.~J. Joshi, F.~Porikli, and N.~Papanikolopoulos.
\newblock Multi-class active learning for image classification.
\newblock In {\em 2009 IEEE Conference on Computer Vision and Pattern
  Recognition}, pages 2372--2379. IEEE, 2009.

\bibitem{joulin2016learning}
A.~Joulin, L.~van~der Maaten, A.~Jabri, and N.~Vasilache.
\newblock Learning visual features from large weakly supervised data.
\newblock In {\em European Conference on Computer Vision}, pages 67--84.
  Springer, 2016.

\bibitem{karpathy2015}
A.~Karpathy and L.~Fei-Fei.
\newblock Deep visual-semantic alignments for generating image descriptions.
\newblock In {\em The IEEE Conference on Computer Vision and Pattern
  Recognition (CVPR)}, June 2015.

\bibitem{karpathy2014}
A.~Karpathy, A.~Joulin, and L.~F. Fei-Fei.
\newblock Deep fragment embeddings for bidirectional image sentence mapping.
\newblock In Z.~Ghahramani, M.~Welling, C.~Cortes, N.~D. Lawrence, and K.~Q.
  Weinberger, editors, {\em Advances in Neural Information Processing Systems
  27}, pages 1889--1897. Curran Associates, Inc., 2014.

\bibitem{kendall2017uncertainties}
A.~Kendall and Y.~Gal.
\newblock What uncertainties do we need in bayesian deep learning for computer
  vision?
\newblock In {\em Advances in neural information processing systems}, pages
  5574--5584, 2017.

\bibitem{kim2016hadamard}
J.-H. Kim, K.-W. On, W.~Lim, J.~Kim, J.-W. Ha, and B.-T. Zhang.
\newblock Hadamard product for low-rank bilinear pooling.
\newblock {\em arXiv preprint arXiv:1610.04325}, 2016.

\bibitem{kingma2014adam}
D.~P. Kingma and J.~Ba.
\newblock Adam: A method for stochastic optimization.
\newblock {\em arXiv preprint arXiv:1412.6980}, 2014.

\bibitem{kiros2014unifying}
R.~Kiros, R.~Salakhutdinov, and R.~S. Zemel.
\newblock Unifying visual-semantic embeddings with multimodal neural language
  models.
\newblock {\em arXiv preprint arXiv:1411.2539}, 2014.

\bibitem{klein2015associating}
B.~Klein, G.~Lev, G.~Sadeh, and L.~Wolf.
\newblock Associating neural word embeddings with deep image representations
  using fisher vectors.
\newblock In {\em Proceedings of the IEEE Conference on Computer Vision and
  Pattern Recognition}, pages 4437--4446, 2015.

\bibitem{kong2014you}
C.~Kong, D.~Lin, M.~Bansal, R.~Urtasun, and S.~Fidler.
\newblock What are you talking about? text-to-image coreference.
\newblock In {\em Proceedings of the IEEE conference on computer vision and
  pattern recognition}, pages 3558--3565, 2014.

\bibitem{krishna2017visual}
R.~Krishna, Y.~Zhu, O.~Groth, J.~Johnson, K.~Hata, J.~Kravitz, S.~Chen,
  Y.~Kalantidis, L.-J. Li, D.~A. Shamma, M.~S. Bernstein, and L.~Fei-Fei.
\newblock Visual genome: Connecting language and vision using crowdsourced
  dense image annotations.
\newblock {\em International Journal of Computer Vision}, 123(1):32--73, 2017.

\bibitem{lewis1994sequential}
D.~D. Lewis and W.~A. Gale.
\newblock A sequential algorithm for training text classifiers.
\newblock In {\em Proceedings of the 17th annual international ACM SIGIR
  conference on Research and development in information retrieval}, pages
  3--12. Springer-Verlag New York, Inc., 1994.

\bibitem{lin2004rouge}
C.-Y. Lin.
\newblock Rouge: A package for automatic evaluation of summaries.
\newblock In {\em Text summarization branches out}, pages 74--81, 2004.

\bibitem{lin2014visual}
D.~Lin, S.~Fidler, C.~Kong, and R.~Urtasun.
\newblock Visual semantic search: Retrieving videos via complex textual
  queries.
\newblock In {\em Proceedings of the IEEE conference on computer vision and
  pattern recognition}, pages 2657--2664, 2014.

\bibitem{lin2014microsoft}
T.-Y. Lin, M.~Maire, S.~Belongie, J.~Hays, P.~Perona, D.~Ramanan,
  P.~Doll{\'a}r, and C.~L. Zitnick.
\newblock Microsoft coco: Common objects in context.
\newblock In {\em European conference on computer vision}, pages 740--755.
  Springer, 2014.

\bibitem{lin2017active}
X.~Lin and D.~Parikh.
\newblock Active learning for visual question answering: An empirical study.
\newblock {\em arXiv preprint arXiv:1711.01732}, 2017.

\bibitem{lu2016hierarchical}
J.~Lu, J.~Yang, D.~Batra, and D.~Parikh.
\newblock Hierarchical question-image co-attention for visual question
  answering.
\newblock In {\em Advances In Neural Information Processing Systems}, pages
  289--297, 2016.

\bibitem{ma2016learning}
L.~Ma, Z.~Lu, and H.~Li.
\newblock Learning to answer questions from image using convolutional neural
  network.
\newblock In {\em Thirtieth AAAI Conference on Artificial Intelligence}, 2016.

\bibitem{mackay1992bayesian}
D.~J. MacKay.
\newblock Bayesian interpolation.
\newblock {\em Neural computation}, 4(3):415--447, 1992.

\bibitem{mairesse2010phrase}
F.~Mairesse, M.~Ga{\v{s}}i{\'c}, F.~Jur{\v{c}}{\'\i}{\v{c}}ek, S.~Keizer,
  B.~Thomson, K.~Yu, and S.~Young.
\newblock Phrase-based statistical language generation using graphical models
  and active learning.
\newblock In {\em Proceedings of the 48th Annual Meeting of the Association for
  Computational Linguistics}, pages 1552--1561. Association for Computational
  Linguistics, 2010.

\bibitem{malinowski2014multi}
M.~Malinowski and M.~Fritz.
\newblock A multi-world approach to question answering about real-world scenes
  based on uncertain input.
\newblock In {\em Advances in neural information processing systems}, pages
  1682--1690, 2014.

\bibitem{malinowski2015ask}
M.~Malinowski, M.~Rohrbach, and M.~Fritz.
\newblock Ask your neurons: A neural-based approach to answering questions
  about images.
\newblock In {\em Proceedings of the IEEE international conference on computer
  vision}, pages 1--9, 2015.

\bibitem{matuszek2012joint}
C.~Matuszek, N.~FitzGerald, L.~Zettlemoyer, L.~Bo, and D.~Fox.
\newblock A joint model of language and perception for grounded attribute
  learning.
\newblock {\em arXiv preprint arXiv:1206.6423}, 2012.

\bibitem{misra2018learning}
I.~Misra, R.~Girshick, R.~Fergus, M.~Hebert, A.~Gupta, and L.~Van Der~Maaten.
\newblock Learning by asking questions.
\newblock In {\em Proceedings of the IEEE Conference on Computer Vision and
  Pattern Recognition}, pages 11--20, 2018.

\bibitem{neal2012bayesian}
R.~M. Neal.
\newblock {\em Bayesian learning for neural networks}, volume 118.
\newblock Springer Science \& Business Media, 2012.

\bibitem{papineni2002bleu}
K.~Papineni, S.~Roukos, T.~Ward, and W.-J. Zhu.
\newblock Bleu: a method for automatic evaluation of machine translation.
\newblock In {\em Proceedings of the 40th annual meeting on association for
  computational linguistics}, pages 311--318. Association for Computational
  Linguistics, 2002.

\bibitem{pennington2014glove}
J.~Pennington, R.~Socher, and C.~Manning.
\newblock Glove: Global vectors for word representation.
\newblock In {\em Proceedings of the 2014 conference on empirical methods in
  natural language processing (EMNLP)}, pages 1532--1543, 2014.

\bibitem{reed2016learning}
S.~Reed, Z.~Akata, H.~Lee, and B.~Schiele.
\newblock Learning deep representations of fine-grained visual descriptions.
\newblock In {\em Proceedings of the IEEE Conference on Computer Vision and
  Pattern Recognition}, pages 49--58, 2016.

\bibitem{ren2015exploring}
M.~Ren, R.~Kiros, and R.~Zemel.
\newblock Exploring models and data for image question answering.
\newblock In {\em Advances in neural information processing systems}, pages
  2953--2961, 2015.

\bibitem{scheffer2001active}
T.~Scheffer, C.~Decomain, and S.~Wrobel.
\newblock Active hidden markov models for information extraction.
\newblock In {\em International Symposium on Intelligent Data Analysis}, pages
  309--318. Springer, 2001.

\bibitem{sener2017active}
O.~Sener and S.~Savarese.
\newblock Active learning for convolutional neural networks: A core-set
  approach.
\newblock {\em arXiv preprint arXiv:1708.00489}, 2017.

\bibitem{settles2012active}
B.~Settles.
\newblock Active learning.
\newblock {\em Synthesis Lectures on Artificial Intelligence and Machine
  Learning}, 6(1):1--114, 2012.

\bibitem{Settles:2008:AAL:1613715.1613855}
B.~Settles and M.~Craven.
\newblock An analysis of active learning strategies for sequence labeling
  tasks.
\newblock In {\em Proceedings of the Conference on Empirical Methods in Natural
  Language Processing}, EMNLP '08, pages 1070--1079, Stroudsburg, PA, USA,
  2008. Association for Computational Linguistics.

\bibitem{settles2008active}
B.~Settles, M.~Craven, and L.~Friedland.
\newblock Active learning with real annotation costs.
\newblock In {\em Proceedings of the NIPS workshop on cost-sensitive learning},
  pages 1--10. Vancouver, CA, 2008.

\bibitem{shen2019learning}
T.~Shen, A.~Kar, and S.~Fidler.
\newblock Learning to caption images through a lifetime by asking questions.
\newblock In {\em Proceedings of the IEEE International Conference on Computer
  Vision}, pages 10393--10402, 2019.

\bibitem{shen2017deep}
Y.~Shen, H.~Yun, Z.~C. Lipton, Y.~Kronrod, and A.~Anandkumar.
\newblock Deep active learning for named entity recognition.
\newblock {\em arXiv preprint arXiv:1707.05928}, 2017.

\bibitem{shih2016look}
K.~J. Shih, S.~Singh, and D.~Hoiem.
\newblock Where to look: Focus regions for visual question answering.
\newblock In {\em Proceedings of the IEEE conference on computer vision and
  pattern recognition}, pages 4613--4621, 2016.

\bibitem{socher2010connecting}
R.~Socher and L.~Fei-Fei.
\newblock Connecting modalities: Semi-supervised segmentation and annotation of
  images using unaligned text corpora.
\newblock In {\em 2010 IEEE Computer Society Conference on Computer Vision and
  Pattern Recognition}, pages 966--973. IEEE, 2010.

\bibitem{sundermeyer2012lstm}
M.~Sundermeyer, R.~Schl{\"u}ter, and H.~Ney.
\newblock Lstm neural networks for language modeling.
\newblock In {\em Thirteenth annual conference of the international speech
  communication association}, 2012.

\bibitem{tong2001support}
S.~Tong and D.~Koller.
\newblock Support vector machine active learning with applications to text
  classification.
\newblock {\em Journal of machine learning research}, 2(Nov):45--66, 2001.

\bibitem{vedantam2015cider}
R.~Vedantam, C.~Lawrence~Zitnick, and D.~Parikh.
\newblock Cider: Consensus-based image description evaluation.
\newblock In {\em Proceedings of the IEEE conference on computer vision and
  pattern recognition}, pages 4566--4575, 2015.

\bibitem{williams1996gaussian}
C.~K. Williams and C.~E. Rasmussen.
\newblock Gaussian processes for regression.
\newblock In {\em Advances in neural information processing systems}, pages
  514--520, 1996.

\bibitem{wu2016ask}
Q.~Wu, P.~Wang, C.~Shen, A.~Dick, and A.~van~den Hengel.
\newblock Ask me anything: Free-form visual question answering based on
  knowledge from external sources.
\newblock In {\em Proceedings of the IEEE Conference on Computer Vision and
  Pattern Recognition}, pages 4622--4630, 2016.

\bibitem{xu2016ask}
H.~Xu and K.~Saenko.
\newblock Ask, attend and answer: Exploring question-guided spatial attention
  for visual question answering.
\newblock In {\em European Conference on Computer Vision}, pages 451--466.
  Springer, 2016.

\bibitem{yang2018visual}
C.-J. Yang, K.~Grauman, and D.~Gurari.
\newblock Visual question answer diversity.
\newblock In {\em Sixth AAAI Conference on Human Computation and
  Crowdsourcing}, 2018.

\bibitem{yang2015multi}
Y.~Yang, Z.~Ma, F.~Nie, X.~Chang, and A.~G. Hauptmann.
\newblock Multi-class active learning by uncertainty sampling with diversity
  maximization.
\newblock {\em International Journal of Computer Vision}, 113(2):113--127,
  2015.

\bibitem{yang2016stacked}
Z.~Yang, X.~He, J.~Gao, L.~Deng, and A.~Smola.
\newblock Stacked attention networks for image question answering.
\newblock In {\em Proceedings of the IEEE conference on computer vision and
  pattern recognition}, pages 21--29, 2016.

\bibitem{zhang2019bertscore}
T.~Zhang, V.~Kishore, F.~Wu, K.~Q. Weinberger, and Y.~Artzi.
\newblock Bertscore: Evaluating text generation with bert.
\newblock {\em arXiv preprint arXiv:1904.09675}, 2019.

\bibitem{zhou2015simple}
B.~Zhou, Y.~Tian, S.~Sukhbaatar, A.~Szlam, and R.~Fergus.
\newblock Simple baseline for visual question answering.
\newblock {\em arXiv preprint arXiv:1512.02167}, 2015.

\bibitem{zhu2016visual7w}
Y.~Zhu, O.~Groth, M.~Bernstein, and L.~Fei-Fei.
\newblock Visual7w: Grounded question answering in images.
\newblock In {\em Proceedings of the IEEE conference on computer vision and
  pattern recognition}, pages 4995--5004, 2016.

\bibitem{zitnick2013learning}
C.~L. Zitnick, D.~Parikh, and L.~Vanderwende.
\newblock Learning the visual interpretation of sentences.
\newblock In {\em Proceedings of the IEEE International Conference on Computer
  Vision}, pages 1681--1688, 2013.

\end{thebibliography}
}
\newpage
\section{Appendix}
We provide more details on how to implement our VQA model. We then motivate our design decisions in developing a new active learning framework for tasks with multiple correct ouputs, by investigating when and how existing uncertainty measurements fail. Next, we explore the active learning framework baselines and provide a more comprehensive quantitative evaluation, as well as a qualitative analysis of our proposed sampling strategy. 

\subsection{Implementation details: VQA model}
We implement a bottom-up, top-down attention~\cite{anderson2018bottom} VQA generation model and use it for all of our active learning experiments. We use a question encoder similar to the LSTM encoder to the Visual-Semantic Encoder. We feed our bottom-up attention mechanism the output hidden state of the LSTM encoder for the question's embedding, along with a $100$ region per corresponding image. We add a dropout layer (with $p=0.5$) to our attention mechanism when we attend over the joint question-image representation and use the original architecture otherwise. We decode the model's answer using an LSTM decoder. We optimize the model with the same hyper-parameters as the Visual-Semantic model and train it for a total of $15$ epochs.

\subsection{Uncertainty estimation with multiple answers: CLEVR experiment}
To motivate our design decisions in developing a new active learning framework for VQA with multiple correct answers, we investigate when and how existing uncertainty measurements fail. To systematically perform this evaluation without confounding factors like noise in real-world datasets, we use the synthetic CLEVR dataset~\cite{johnson2017clevr}. Even though CLEVR only has questions with one correct answer, we modify the answers by introducing paraphrases. Our insights from experimenting on a modified-CLEVR motivate both the importance of moving to different uncertainty estimation for language generation models, as well as the specific uncertainty measure described in the next section.

\paragraph{Modifying CLEVR to include paraphrases}
CLEVR is a diagnostic dataset that tests a range of visual reasoning abilities, including Visual Question Answering. It contains minimal biases and has detailed annotations describing the kind of reasoning each question requires. The dataset is composed of a train set of $700k$ $(I,Q,A)$ triples with multiple answer categories including binary (yes / no), attributes, counts, objects and spatial relationships.

In a real world dataset, some answers will have multiple correct answers while other answers might only have one. To mimic such a setup, we modify CLEVR by taking a fraction of answer categories and replacing them with synonyms. For example, we corrupt the answer `yes' to seven different tokens $\xrightarrow{} \{$`yes', `yeah', $\ldots$, `yup'$\}$. Specifically, for every answer that is `yes', we randomly modify it to one of the seven paraphrases. This is a conservative modification; in language, there are usually more than seven ways of expressing the same meaning.

\begin{figure}[t]
\begin{center}
\includegraphics[width=\columnwidth]{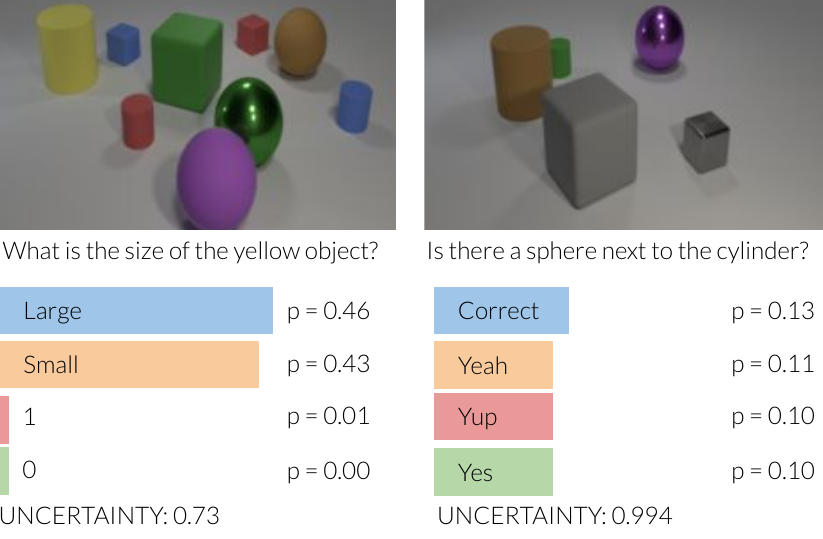}
\caption{Even though the model is uncertain about the question on the \textbf{left}, existing uncertainty measurements assign it a lower uncertainty than the example on the \textbf{right}. When an input has multiple correct answers, existing measures overestimate uncertainty as they are unable to relate which outputs are paraphrases.}
\label{fig:clevr}
\end{center}
\end{figure}

\paragraph{Overestimation of uncertainty}
We train a state-of-the art VQA model~\cite{yang2016stacked}, on both the modified-CLEVR dataset for $10$ epochs with a learning rate of $4e-3$. Next, for all the data points in modified-CLEVR's validation set, we measure the model's uncertainty from the model's outputs and compare how uncertainty measurements differ for questions with multiple correct answers and those with only one. 

We report uncertainty scores using entropy, which is the most popular uncertainty measurement used in active learning settings. We run similar experiments with other measures, like least confidence and margin, but omit them from our analysis here as they follow the same trend. Since exactly measuring entropy is intractable, we approximate entropy by decoding the answer with a beam size of $5$~\cite{shen2017deep}.

Qualitatively, Figure~\ref{fig:clevr} demonstrates how the model's uncertainty scores differ for questions with one or multiple correct answers. On the left, we show an example of a question that the model is relatively unsure about. It assigns most of its weight to the correct answer (``large'') but also assigns a sizable weight to an incorrect answer (``small''). However, the uncertainty associated with this question is lower than the example we show on the right. When multiple correct answers exist, the model fails to choose a clear answer and tends to assign similar weights to multiple answers with the same meaning. Even though we can interpret such a result as the model learning that for such a question, multiple correct candidates exist, uncertainty measures are unaware of which answers are paraphrases and overestimate uncertainty. This scenario showcases a common failure case of active learning with existing uncertainty measurements, which would choose to collect more labels for the example with multiple correct answers, even though the model might benefit more from sampling the question on the left.

\begin{figure}[t]
\begin{center}
\includegraphics[width=\columnwidth]{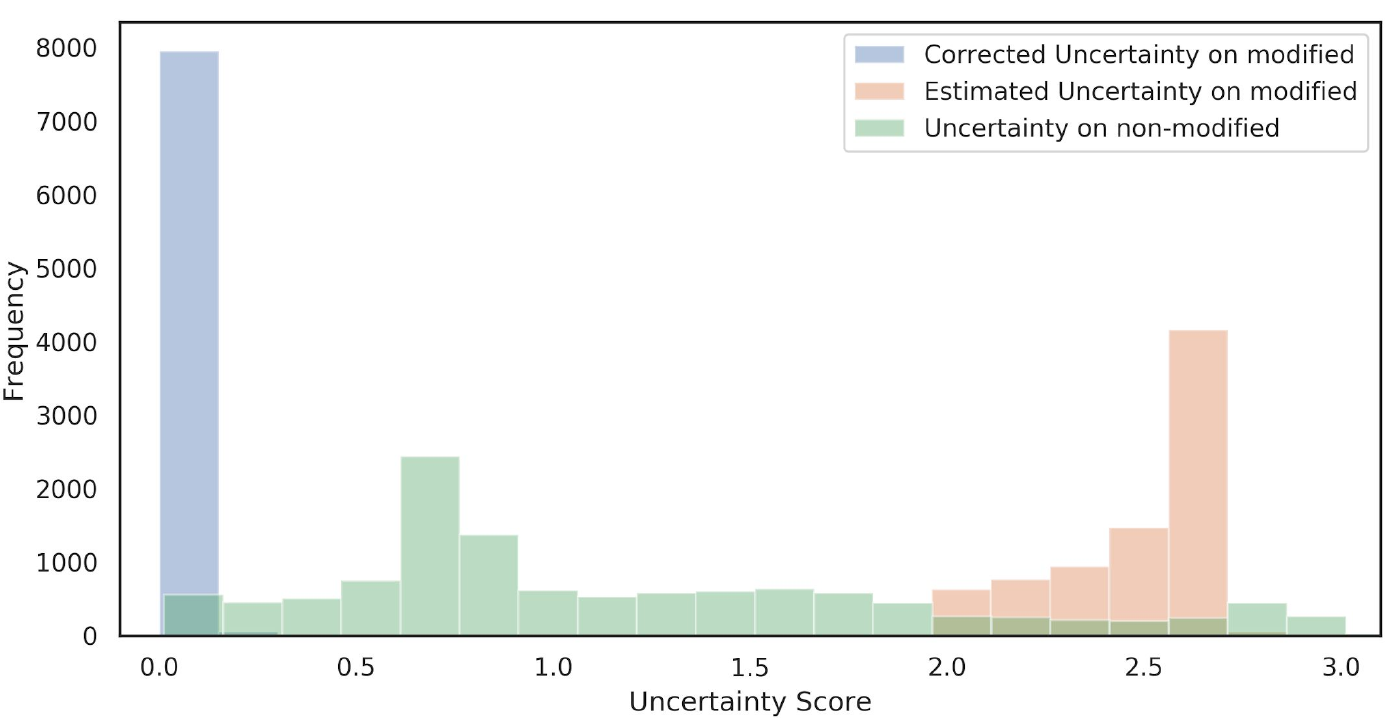}
\caption{As existing uncertainty measurements tend to overestimate uncertainty when an input has multiple correct answers, we end up sampling data points that the model already knows how to answer reducing active learning efficiency.}
\label{fig:clevr_exp}
\end{center}
\end{figure}

Quantitatively, we show how prevalent this problem is for all questions in the modified-CLEVR's validation set in Figure~\ref{fig:clevr_exp}. Entropy assigns high uncertainty for all questions with multiple correct answers (shown in orange). And majority of these questions are assigned a higher uncertainty score than questions with a single correct answer (shown in green). Finally, if we correct the uncertainty scores by summing up the weights assigned to all the synonyms together, we find that the model is actually quite certain about a lot of these questions (shown in blue).

Our experiments with CLEVR allows us to conclude that we need an uncertainty measurement that can perform a similar correction, moving overestimated uncertainty scores (as shown in orange) to their correct values (as shown in blue). We need to design a mechanism which suggests which answers are semantically similar so that such a correction can be performed.

\subsection{Active learning framework}

We visualize the active learning framework described in the main paper using an algorithm in Algorithm~\ref{alg:active_learning}. We also add additional notes and details about the baselines used below:

\begin{algorithm}[t!]
\caption{Pool Based Active learning for VQA}
\label{alg:active_learning}
\begin{algorithmic}[1]
  \scriptsize
  \STATE Initialize an initial training set $D_{train}$
  \STATE Initialize VQA model and pretrain it on $D_{train}$ to get $f^0$
  \FOR{$t$ in $1$ to $T$}
  \STATE Get a new pool $P^t$ of $N$ pairs of $(Q,I)$
  \STATE initialize a list $S$
  \FOR{each pair $(Q, I) \in  P$}
    \STATE Measure uncertainty $s_{qi} \gets U(Q, I)$,
    \STATE Add uncertainty score to list $S \gets S + s_{qi}$
  \ENDFOR
  \STATE SAMPLES $ \gets \textrm{argsort}(S)[::-1][:K]$
  \STATE Annotate SAMPLES using an oracle
  \STATE Add new data to training set $D_{train}\gets D_{train} +$ SAMPLES
  \STATE Retrain $f^t$ using the updated $D$ to get $f^{t+1}$
  \ENDFOR
\end{algorithmic}
\end{algorithm}

\paragraph{Random Sampling (Passive Learning):} We choose $U$ following a uniformly random distribution from $P^{'}$.
$$U = unif(P^{'}), $$ where $unif()$ returns k data points from $P^{'}$ following a uniform distribution.

\paragraph{Least Confidence Sampling  \cite{culotta2005reducing}:} This approach queries the instances for which our model has the least conﬁdence in its most likely generated sequence. We choose $U$ as the set with the top $K$ points from $P^{'}$ that the current VQA model has the least confidence in generating an answer for. We define our acquisition function as follows:
$$s(Q, I) = 1 - P(a^*| Q, I),$$
where $P(a^*| Q, I)$ is the probability of the most confident model response.

\paragraph{Margin Sampling \cite{scheffer2001active}:} This approach queries the instances with the least margin between the output probabilities for the two most likely generated answers of our model. In this strategy, we use a beam size of 2 and choose $U$ as the set with the top K points from $P^{’}$ with the smallest margin. We define our acquisition function as follows:
$$s(Q, I) = 1 + P(a_2| Q, I) - P(a_1 | Q, I),$$
where $P(a_i| Q, I)$ is the probability of the $i^{th}$ most confident model response, $i \in \{1, 2\}$.

\paragraph{Maximum Entropy Sampling \cite{shen2017deep}:} This approach queries the instances that maximize the entropy of our models output. We choose $U$ as the $K$ data points with the highest prediction entropy. We define our acquisition function as follows:
$$s(Q, I) = - \sum_{i=1}^{b} P_i(a|Q, I) logP_i(a|Q, I),$$
In order to measure entropy of our predictions in a VQA setting, we use a beam size of $b$ to get $b$ different predictions with probabilities  $P_i(a|Q, I)$ for  $i =1, 2, .., b$.

\subsection{More active learning performance:}
The experiments we ran in the main paper can also be evaluated using additional language metrics like BLEU, ROUGE and BERT F1. We include those metrics here. They follow the same trend as the other metrics. We also report model accuracy. 

Results on Visual Genome are shown in Table~\ref{tab:vgcomp} and results on VQA 2.0 are shown in Table~\ref{tab:vqacomp}.

\begin{table*}[h]
    \begin{center}\small
		\setlength{\tabcolsep}{.25em}
		\begin{tabular}{l|cccccc|cccccc|cccccc}
			\midrule
			VG & \multicolumn{5}{c}{Bert F1} &  & \multicolumn{5}{c}{Bert Recall} &  & \multicolumn{5}{c}{Bert Precision}\\
			\midrule
			AL iteration &  1 & 2 & 3 & 4 & 5 &    &  1 & 2 & 3 & 4 & 5 &     &  1 & 2 & 3 & 4 & 5\\ 
			\midrule
			\midrule
			Dataset Size (\%)&  10 & 15 & 20 & 25 & 30 &    &  10 & 15 & 20 & 25 & 30 &     &  10 & 15 & 20 & 25 & 30\\ 
			\midrule
			\texttt{Random}& 66.12 & 66.75 & 67.16 & 67.46 & 67.74 & & 66.96 & 67.63 & 68.05 & 68.38 & 68.60 & & 65.31 & 65.88 & 66.30 & 66.57 & 66.89 \\
			\texttt{Margin}& 65.92 & 66.84 & 67.23 & 67.60 & 67.87 & & 66.83 & 67.65 & 68.10 & 68.51 & 68.77 & & 65.04 & 66.04 & 66.38 & 66.71 & 66.98 \\
			\texttt{LC}& 66.17 &  66.82 &  67.36 &  67.59 &  67.88 & & 67.05 & 67.72 &  68.18 & 68.55 & 68.79 & & 65.28 & 65.95 & 66.45 &  66.65 & 66.99 \\
			\texttt{Max Entropy}& 66.14 &  66.84 &  67.36 &  67.68 &  67.86 & & 67.00 & 67.69 & 68.26 & 68.62 & 68.74& & 65.30 & 66.01 & 66.48 & 66.77 & 67.00\\
			\texttt{Ours} &   \textbf{68.10} &  \textbf{68.94} & \textbf{69.47} &  \textbf{69.81} &  \textbf{70.14} & & \textbf{69.74} & \textbf{70.51} & \textbf{70.99} & \textbf{71.35} & \textbf{71.63} & & \textbf{66.54} &  \textbf{67.44} & \textbf{68.01} & \textbf{68.34} & \textbf{68.70} \\ 
			\midrule
		\end{tabular}
	\end{center}
    \begin{center}\small
		\setlength{\tabcolsep}{.25em}
		\scalebox{0.95}{
		\begin{tabular}{l|cccccc|cccccc|cccccc}
			\midrule
			VG & \multicolumn{5}{c}{METEOR} &  & \multicolumn{5}{c}{CIDEr} & & \multicolumn{5}{c}{ROUGE-L} \\
			\midrule
			AL iteration &  1 & 2 & 3 & 4 & 5 &    &  1 & 2 & 3 & 4 & 5 &     &  1 & 2 & 3 & 4 & 5\\ 
			\midrule
			\midrule
			Dataset Size (\%)&  10 & 15 & 20 & 25 & 30 &    &  10 & 15 & 20 & 25 & 30 &     &  10 & 15 & 20 & 25 & 30\\ 
			\midrule
			\texttt{Random}& 12.96 &  13.57 &  14.00 &  14.37 &  14.69 & & 87.01 &  91.75 &  94.90 &  97.63 & 100.09 & & 30.38 & 31.67 &  32.55 & 33.31 & 33.96\\
			\texttt{Margin}& 12.82 & 13.70 & 14.11 & 14.52 & 14.70 & &86.10 & 92.28 & 95.66 &  98.70 & 100.30 & & 30.06 &  31.68 &  32.57 &  33.37 &  33.79\\
			\texttt{LC}& 12.91 & 13.62 & 14.16 &  14.43 &  14.76 & &86.76 & 91.80 & 95.91 & 97.94 & 100.22 & & 30.17 & 31.44 & 32.53 & 33.06 & 33.65 \\
			\texttt{Max Entropy}& 12.92 &  13.66 & 14.22 & 14.49 & 14.76 & & 87.05 & 92.26 & 96.36 & 98.47 & 100.45& & 30.31 & 31.66 & 32.77 & 33.36 & 33.89\\
			\texttt{Ours} &  \textbf{14.66} &  \textbf{15.53} &  \textbf{16.12} & \textbf{16.47} &  \textbf{16.91}& &  \textbf{102.19} & \textbf{108.87} & \textbf{112.80} & \textbf{115.67} &  \textbf{118.67} & &\textbf{35.02} &  \textbf{36.70} &  \textbf{37.62} & \textbf{38.32} &  \textbf{39.02}\\
			\midrule 
		\end{tabular}
		}
	\end{center}
	\begin{center}\small
		\setlength{\tabcolsep}{.25em}
		\begin{tabular}{l|cccccc|cccccc|cccccc}
			\midrule
			VG & \multicolumn{5}{c}{Bleu-1} &  & \multicolumn{5}{c}{Bleu-2} & & \multicolumn{5}{c}{Bleu-3} \\
			\midrule
			AL iteration &  1 & 2 & 3 & 4 & 5 &    &  1 & 2 & 3 & 4 & 5 &     &  1 & 2 & 3 & 4 & 5\\ 
			\midrule
			\midrule
			Dataset Size (\%)&  10 & 15 & 20 & 25 & 30 &    &  10 & 15 & 20 & 25 & 30 &     &  10 & 15 & 20 & 25 & 30\\ 
			\midrule
			\texttt{Random}& 29.1 & 29.96 & 30.52 & 31.12 & 31.74 & & 18.86 & 19.63 & 20.08 & 20.57 & 20.94 & &12.72 & 13.41 & 13.81 & 14.2 & 14.47\\
			\texttt{Margin}& 28.7 & 30.22 & 30.85 & 31.41 & 31.63 & &18.71 & 19.91 & 20.47 & 21.03 & 21.21 & & 12.67 & 13.62 & 14.17 & 14.72 & 14.92\\
			\texttt{LC}& 28.98 & 29.94 & 30.96 & 31.14 & 31.77 & & 18.99 & 19.78 & 20.59 & 20.95 & 21.31 & & 12.92 & 13.66 & 14.26 & 14.71 & 14.98 \\
			\texttt{Max Entropy}& 29.0 & 29.99 & 31.01 & 31.25 & 31.79 & & 18.91 & 19.75 & 20.56 & 20.84 & 21.17 & & 12.81 & 13.52 & 14.17 & 14.51 & 14.72\\
			\texttt{Ours} &  \textbf{31.59} & \textbf{33.07} & \textbf{33.75} & \textbf{34.29} & \textbf{35.19} & & \textbf{23.16} & \textbf{24.22} & \textbf{24.82} & \textbf{25.41} & \textbf{25.91} & & \textbf{17.21} & \textbf{18.17} & \textbf{18.67} & \textbf{19.37} & \textbf{19.75}\\
			\midrule 
		\end{tabular}
	\end{center}
	\begin{center}\small
		\setlength{\tabcolsep}{.25em}
		\begin{tabular}{l|cccccc|cccccc}
			\midrule
			VG & \multicolumn{5}{c}{Bleu-4} &  & \multicolumn{5}{c}{Accuracy}\\
			\midrule
			AL iteration &  1 & 2 & 3 & 4 & 5 &    &  1 & 2 & 3 & 4 & 5\\ 
			\midrule
			\midrule
			Dataset Size (\%)&  10 & 15 & 20 & 25 & 30 &    &  10 & 15 & 20 & 25 & 30\\ 
			\midrule
			\texttt{Random}& 7.94 & 8.40 & 8.88 & 9.15 & 9.26 &  & 16.36 & 17.40 & 18.12 & 18.7 & 19.21\\
			\texttt{Margin}& 7.98 & 8.72 & 9.20 & 9.71 & 9.93 & &16.14 & 17.34 & 18.05 & 18.67 & 19.02\\
			\texttt{LC}& 8.17 & 8.85 & 9.23 & 9.67 & 9.88 & &16.14 & 17.11 & 17.94 & 18.40 & 18.81 \\
			\texttt{Max Entropy}& 8.05 & 8.74 & 9.10 & 9.51 & 9.47 & & 16.32 & 17.43 & 18.19 & 18.69 & 19.09\\
			\texttt{Ours} &  \textbf{13.20}& \textbf{14.04} & \textbf{14.16} & \textbf{15.18} & \textbf{15.40}& &  \textbf{19.85} & \textbf{21.25} & \textbf{22.07} & \textbf{22.60}& \textbf{23.07}\\
			\midrule 
		\end{tabular}
	\end{center}
	\caption{We report our model’s efficacy with multiple metrics. We use language modeling metrics to measure its capability to generate answers similar to the ground truth as we progress through the AL process. Scores are multiplied by 100 to show more significant digits.}
	\label{tab:vgcomp}
\end{table*}

\begin{table*}[h]
    \begin{center}\small
		\setlength{\tabcolsep}{.25em}
		\begin{tabular}{l|cccccc|cccccc|cccccc}
			\midrule
			VQA 2.0 & \multicolumn{5}{c}{Bert F1} &  & \multicolumn{5}{c}{Bert Recall} &  & \multicolumn{5}{c}{Bert Precision}\\
			\midrule
			AL iteration &  1 & 2 & 3 & 4 & 5 &    &  1 & 2 & 3 & 4 & 5 &     &  1 & 2 & 3 & 4 & 5\\ 
			\midrule
			\midrule
			Dataset Size (\%)&  10 & 15 & 20 & 25 & 30 &    &  10 & 15 & 20 & 25 & 30 &     &  10 & 15 & 20 & 25 & 30\\ 
			\midrule
			\texttt{Random}& 83.82 & 84.87 & 85.57 & 86.04 & 86.4 & & 83.83 & 84.9 & 85.58 & 86.05 & 86.42 & & 83.8 & 84.85 & 85.57 & 86.02 & 86.39 \\ 
			\texttt{Margin}& 83.68 & 84.92 & 85.32 & 85.8 & 86.29 & & 83.7 & 84.93 & 85.33 & 85.82 & 86.31 & & 83.66 & 84.91 & 85.32 & 85.78 & 86.28 \\
			\texttt{LC}& 84.08 & 85.19 & 85.71 & 86.1 & 86.03 & & 84.1 & 85.20 & 85.73 & 86.12 & 86.04 & & 84.06 & 85.17 & 85.69 & 86.09 & 86.02 \\
			\texttt{Max Entropy}& 84.30 & 85.05 & 85.54 & 84.8 & 86.42 & & 84.32 & 85.07 & 85.56 & 84.83 & 86.44 & & 84.29 & 85.03 & 85.52 & 84.77 & 86.39\\
			\texttt{Ours} &   \textbf{85.02} & \textbf{85.87} & \textbf{86.44} & \textbf{86.85} & \textbf{87.01} & & \textbf{85.07} & \textbf{85.88} & \textbf{86.47} & \textbf{86.89} & \textbf{87.02} & & \textbf{84.98} & \textbf{85.85} & \textbf{86.42} & \textbf{86.82} & \textbf{87.00} \\
			
			\midrule
		\end{tabular}
	\end{center}
    \begin{center}\small
		\setlength{\tabcolsep}{.25em}
		\scalebox{0.95}{
		\begin{tabular}{l|cccccc|cccccc|cccccc}
			\midrule
			VQA 2.0& \multicolumn{5}{c}{METEOR} &  & \multicolumn{5}{c}{CIDEr} & & \multicolumn{5}{c}{ROUGE-L} \\
			\midrule
			AL iteration &  1 & 2 & 3 & 4 & 5 &    &  1 & 2 & 3 & 4 & 5 &     &  1 & 2 & 3 & 4 & 5\\ 
			\midrule
			\midrule
			Dataset Size (\%)&  10 & 15 & 20 & 25 & 30 &    &  10 & 15 & 20 & 25 & 30 &     &  10 & 15 & 20 & 25 & 30\\ 
			\midrule
			\texttt{Random}& 29.00 & 32.74 & 35.42 & 35.92 & 36.73 & & 112.51 & 120.92 & 126.53 & 130.08 & 133.25 & & 43.96 & 47.22 & 49.31 & 50.69 & 51.94\\
			\texttt{Margin}& 28.60 & 32.88 & 32.84 & 35.65 & 36.66 & &111.04 & 120.35 & 123.17 & 127.08 & 131.65 & & 43.4 & 46.96 & 47.98 & 49.53 & 51.31\\
			\texttt{LC}& 29.49 & 34.21 & 34.92 & 35.31 & 35.85 & & 113.35 & 121.95 & 125.72 & 128.98 & 127.84 & & 44.25 & 47.53 & 48.97 & 50.21 & 49.72 \\
			\texttt{Max Entropy}& 30.06 & 32.72 & 34.75 & 33.90 & 37.03 & & 114.96 & 119.50 & 123.54 & 119.54 & 130.58 & & 44.84 & 46.56 & 48.05 & 46.62 & 50.80\\
			\texttt{Ours} &  \textbf{31.21} & \textbf{34.58} & \textbf{36.82} & \textbf{37.62} & \textbf{37.03} & & \textbf{118.64} & \textbf{125.72} & \textbf{130.23} & \textbf{134.07} & \textbf{136.36} & & \textbf{46.35} & \textbf{48.96} & \textbf{50.67} & \textbf{52.17} & \textbf{52.98}\\
			\midrule 
		\end{tabular}
		}
	\end{center}
	\begin{center}\small
		\setlength{\tabcolsep}{.25em}
		\begin{tabular}{l|cccccc|cccccc|cccccc}
			\midrule
			VG & \multicolumn{5}{c}{Bleu-1} &  & \multicolumn{5}{c}{Bleu-2} & & \multicolumn{5}{c}{Bleu-3} \\
			\midrule
			AL iteration &  1 & 2 & 3 & 4 & 5 &    &  1 & 2 & 3 & 4 & 5 &     &  1 & 2 & 3 & 4 & 5\\ 
			\midrule
			\midrule
			Dataset Size (\%)&  10 & 15 & 20 & 25 & 30 &    &  10 & 15 & 20 & 25 & 30 &     &  10 & 15 & 20 & 25 & 30\\ 
			\midrule
			\texttt{Random}& 43.93 & 47.18 & 49.35 & 50.74 & 51.94 &  & 38.7 & 41.98 & 44.29 & 45.17 & 46.81 &  &  37.55 & 40.7 & 42.8 & 42.89 & 45.05\\
			\texttt{Margin}& 43.38 & 46.96 & 48.06 & 49.56 & 51.34 &  & 37.35 & 41.4 & 42.75 & 44.77 & 45.58 &  & 35.42 & 40.94 & 41.08 & 43.96 & 43.31\\
			\texttt{LC}& 44.30 & 47.60 & 49.05 & 50.30 & 49.85 &  & 38.92 & 42.57 & 44.04 & 45.53 & 45.58 &  & 35.79 & 40.15 & 41.73 & 43.92 & 43.03 \\
			\texttt{Max Entropy}& 44.87 & 46.65 & 48.21 & 46.61 & 50.89 &  & 40.79 & 42.41 & 45.12 & 40.96 & 48.40 &  & 40.26 & 40.52 & 44.06 & 35.75 & 48.58\\
			\texttt{Ours} & \textbf{46.33} & \textbf{49.07} & \textbf{50.78} & \textbf{52.26} & \textbf{53.15} &  & \textbf{42.52} & \textbf{44.79} & \textbf{47.31} & \textbf{49.09} & \textbf{48.97} &  & \textbf{42.55} & \textbf{43.38} & \textbf{46.99} & \textbf{48.81} & \textbf{47.05}\\
			\midrule 
		\end{tabular}
	\end{center}
	\begin{center}\small
		\setlength{\tabcolsep}{.25em}
		\begin{tabular}{l|cccccc|cccccc}
			\midrule
			VG & \multicolumn{5}{c}{Bleu-4} &  & \multicolumn{5}{c}{Accuracy}\\
			\midrule
			AL iteration &  1 & 2 & 3 & 4 & 5 &    &  1 & 2 & 3 & 4 & 5\\ 
			\midrule
			\midrule
			Dataset Size (\%)&  10 & 15 & 20 & 25 & 30 &    &  10 & 15 & 20 & 25 & 30\\ 
			\midrule
			\texttt{Random}& \textbf{16.07} & 19.59 & 6.92 & 13.09 & 11.53 & & 43.36 & 46.6 & 48.69 & 50.05 & 51.32 \\
			\texttt{Margin}& 0.01 & 10.64 & 10.63 & 7.50 & 18.92 & & 42.77 & 46.33 & 47.35 & 48.90 & 50.66\\
			\texttt{LC}&5.18 & 4.36 & 4.81 & \textbf{13.10} & 14.64 & & 43.60 & 46.88 & 48.32 & 49.56 & 49.05 \\
			\texttt{Max Entropy}& 0.33 & 18.00 & 18.46 & 11.94 & 26.87 & & 44.22 & 45.91 & 47.41 & 46.00 & 50.19\\
			\texttt{Ours} & 8.85 & \textbf{19.83} & \textbf{33.98} & 12.88 & \textbf{27.01} & & \textbf{45.68} & \textbf{48.27} & \textbf{50.03} & \textbf{51.52} & \textbf{52.31}\\
			\midrule 
		\end{tabular}
	\end{center}
	\caption{We report our model’s efficacy with multiple metrics. We use language modeling metrics to measure its capability to generate answers similar to the ground truth as we progress through the AL process. Scores are multiplied by 100 to show more significant digits.}
	\label{tab:vqacomp}
\end{table*}

\subsection{Qualitative analysis between \texttt{Margin} vs \texttt{Ours}}
As discussed in section 5.2, we qualitatively study the behavior of both \texttt{Margin}, as as our solution on different types of questions. We find that our solution out-performs \texttt{Margin} in all types of questions by a considerable margin. We also study the percentage of sampled points with respect to performance for each type of question. We find that our solution not only performs better when we sample more points but also in cases where we sample less. 

\begin{figure}[t]
\begin{center}
\includegraphics[width=\linewidth]{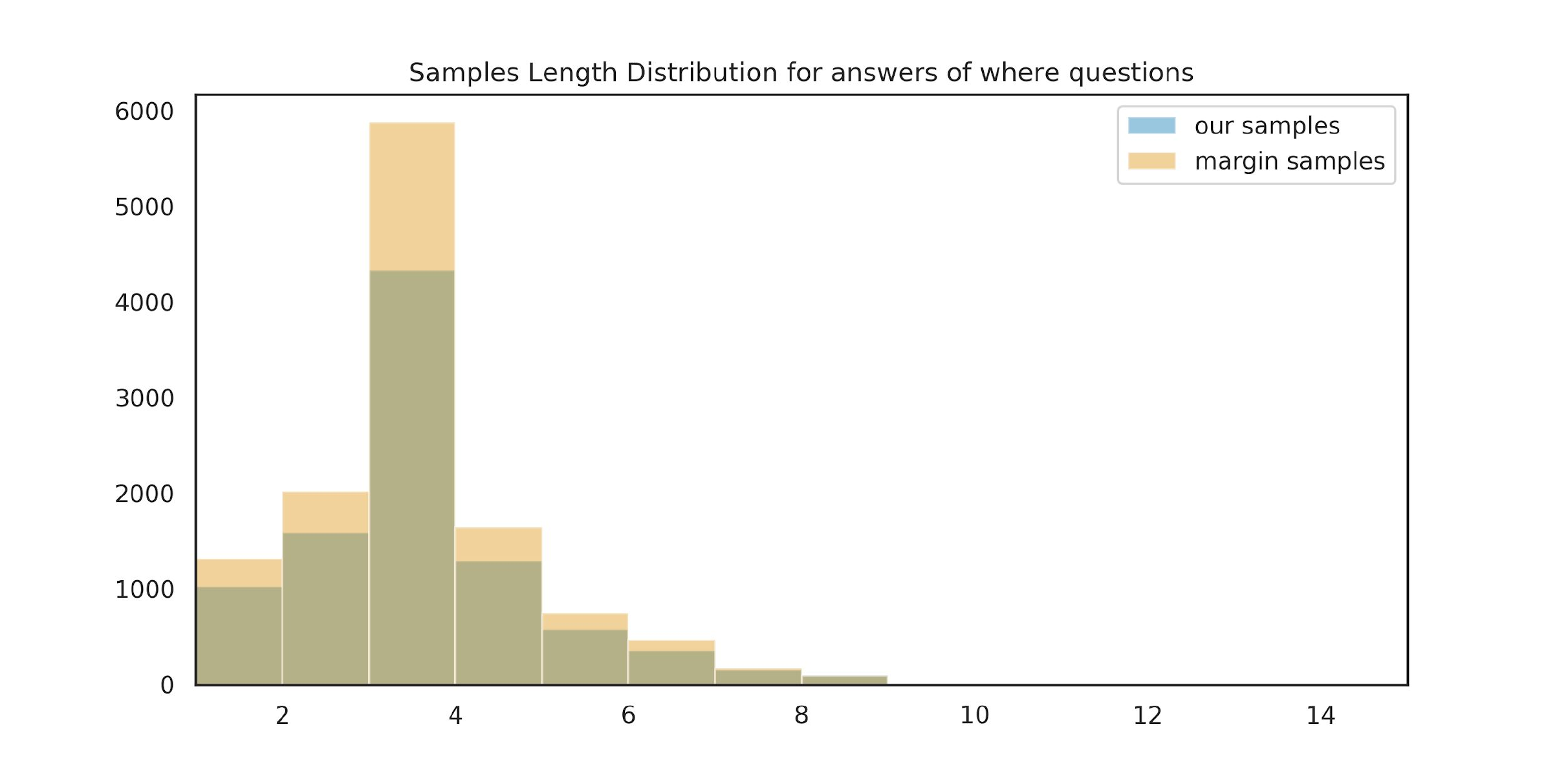}
\caption{Distribution of sampled answer lengths in ``where'' questions for \texttt{Ours} and \texttt{Margin} in the 3rd iteration.}
\label{fig:where}
\end{center}
\end{figure}

\begin{figure}[t]
\begin{center}
\includegraphics[width=\linewidth]{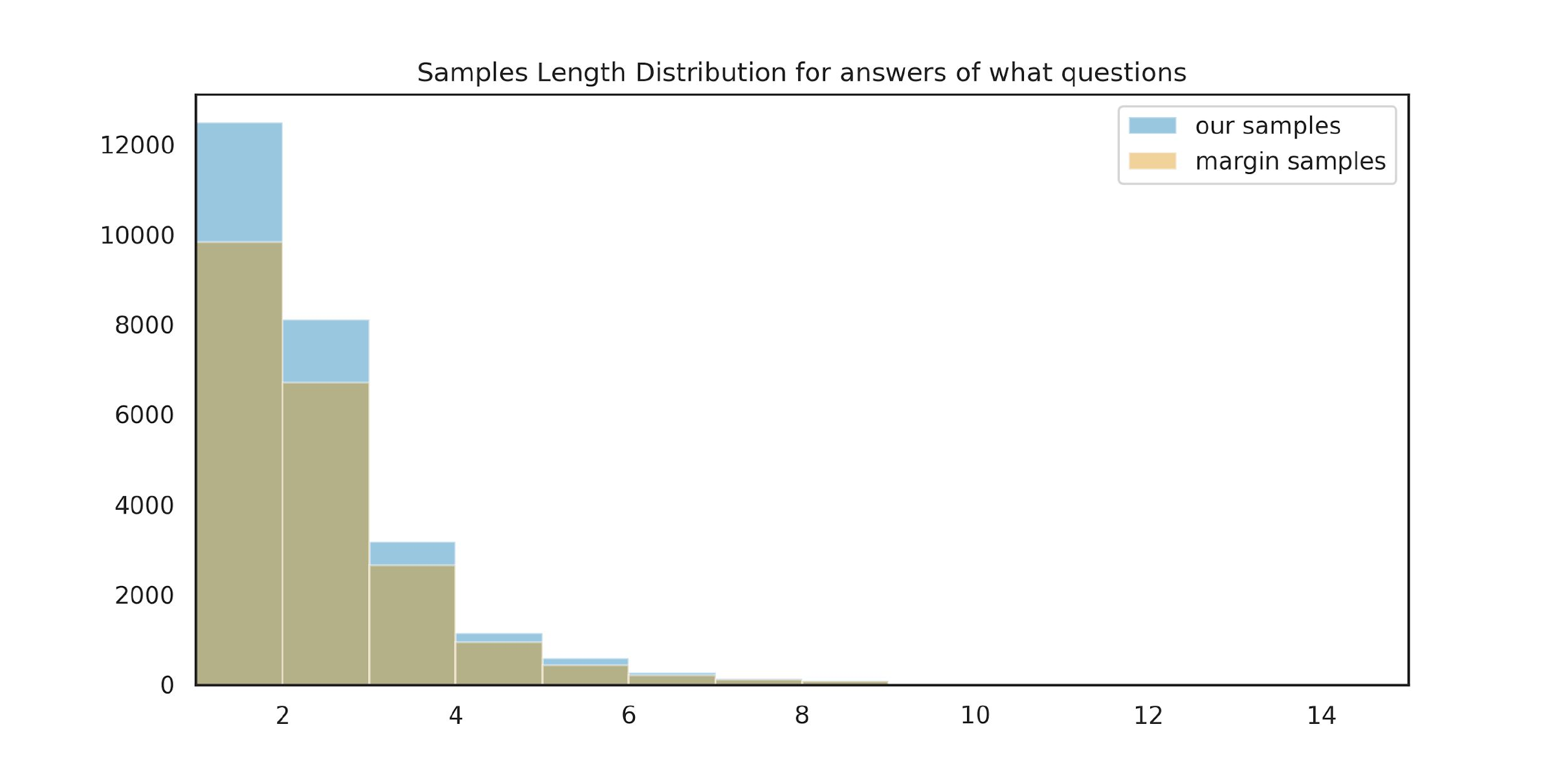}
\caption{Distribution of sampled answer lengths in ``what'' questions for \texttt{Ours} and \texttt{Margin} in the 3rd iteration.}
\label{fig:what}
\end{center}
\end{figure}

In order to understand this behavior, we study samples from the $3^{rd}$ iteration of active learning, where we sample significantly less ``where'' questions and more ``what'' questions and yet perform better than \texttt{Margin} on both types. We investigate the distribution of answer length for samples in both types of questions. Figure~\ref{fig:where} and~\ref{fig:what} show the sampled distributions. We find that \texttt{Margin} samples $25.3\%$ more answers of length $\geq 4$ on ``where'' questions ($3271$ vs $2647$).

We also look at the average length of answers in Visual Genome~\cite{krishna2017visual} and find that ``What'' questions have an average length of $1.9(\pm1.2)$ while ``where'' questions have an average length of $3.1(\pm1.5)$.  Longer answers typically have a higher potential of having paraphrases. We find this assumption to be true qualitatively: an example of the redundancy of \texttt{Margin} is shown in Table~\ref{tab:margin-q}. We look at sampled points containing the word ``building''. We find that \texttt{Margin} heavily re-samples concepts that are already available or have paraphrases in the training set. We also report the corresponding samples for \texttt{Ours} in Table~\ref{tab:ours-q} and find that we sample less paraphrases and are less redundant.

\begin{table*}[]
\begin{center}\small
    \begin{tabular}{|l|l|}
    \hline
    \texttt{Margin} $3^{rd}$ iteration samples & Paraphrases already in the Training Set Points\\
    \hline
    \begin{tabular}[c]{@{}l@{}}between the buildings \\ between the red brick buildings \\ over the street and between buildings \end{tabular}                                                                            & \begin{tabular}[c]{@{}l@{}}between the buildings\\ in between two buildings\\ between the pear and lemon buildings\\ overhead between the buildings\\ between two buildings\\ between buildings\\ hanging between buildings\\ in between buildings\end{tabular}                   \\
    \hline
    \begin{tabular}[c]{@{}l@{}}above the buildings \\ on buildings \\ on the buildings \\ on the buildings behind the vehicles \end{tabular}                                                                                & \begin{tabular}[c]{@{}l@{}}above buildings\\ on buildings \\ above the train and buildings\\ on the buildings \\ on top of buildings\\ on the buildings in the background\\ on the buildings\\ above the city buildings\\ above the buildings \\ hanging over buildings\end{tabular} \\
    \hline
    \begin{tabular}[c]{@{}l@{}}in front of the buildings \\ in front of buildings \\ parked in front of the buildings \end{tabular}                                                                                        & \begin{tabular}[c]{@{}l@{}}in front of the buildings and tower\\ in front of buildings \\ in front of the buildings \end{tabular}\\
    \hline
    \begin{tabular}[c]{@{}l@{}}near the buildings \\ near buildings \\ next to the large brick buildings \\ by the buildings \\ in the street , near buildings \\ next to the buildings \\ by the tall buildings \end{tabular} & \begin{tabular}[c]{@{}l@{}}near the buildings\\ beside the buildings\\ by the trees and buildings\\ near buildings\\ close to buildings\\ next to the buildings\end{tabular}\\
    \hline
    \begin{tabular}[c]{@{}l@{}} behind buildings \\ behind the buildings  \end{tabular}& \begin{tabular}[c]{@{}l@{}}behind the buildings \\ behind the other buildings \\ behind all the buildings \\
    \end{tabular} \\
    \hline
\end{tabular}   
\end{center}
\caption{We look at all the sampled phrases containing the word ``building'' with \texttt{Margin}. We realize that we end up heavily re-sampling concepts that are already available in our training set. This shows the redundancy of this strategy and explains why sampling more points leads to worse performance.}
\label{tab:margin-q}
\end{table*}

\begin{table*}[]
\begin{center}\small
    \begin{tabular}{|l|l|}
    \hline
    \texttt{Ours} $3^{rd}$ iteration samples & Paraphrases already in the Training Set Points\\
    \hline
    \begin{tabular}[c]{@{}l@{}}between the buildings \\ between the two buildings \end{tabular}                                                                            & \begin{tabular}[c]{@{}l@{}}between two buildings \\in the middle of the buildings \\ overhead between the buildings \\ over the street and between buildings \\ between the buildings \\ in between the buildings \\ in between buildings \end{tabular}\\
    \hline
    \begin{tabular}[c]{@{}l@{}}above the buildings \\ on buildings \end{tabular} & \begin{tabular}[c]{@{}l@{}}on top of buildings \\ on the buildings  \\ on buildings  \\ above the buildings\end{tabular} \\
    \hline
    \begin{tabular}[c]{@{}l@{}}in front of buildings \end{tabular} & \begin{tabular}[c]{@{}l@{}}in front of the buildings and tower \\ in front of buildings \\ in front of the buildings \end{tabular}\\
    \hline
    \begin{tabular}[c]{@{}l@{}}near buildings \\ by the buildings \end{tabular} & \begin{tabular}[c]{@{}l@{}}near the buildings \\ by the trees and buildings \\ near buildings \\ sidewalk along buildings \\ near the brown buildings \\ by the buildings\end{tabular}\\
    \hline
    \begin{tabular}[c]{@{}l@{}} behind older buildings \\ behind the buildings  \end{tabular}& \begin{tabular}[c]{@{}l@{}}behind the other buildings \\ behind the buildings 
    \end{tabular} \\ 
    \hline
\end{tabular}   
\end{center}
\caption{We look at all the sampled phrases containing the word ``building'' with \texttt{Margin}. We find that compared to \texttt{Margin}, we cover the same range of concepts while being less redundant.}
\label{tab:ours-q}
\end{table*}

\end{document}